\documentclass{article}

\usepackage{multirow}
\usepackage{adjustbox}
\usepackage[preprint]{neurips_2026}

\usepackage[utf8]{inputenc} 
\usepackage[T1]{fontenc}    
\usepackage{hyperref}       
\usepackage{url}            
\usepackage{booktabs}       
\usepackage{amsfonts}       
\usepackage{nicefrac}       
\usepackage{microtype}      
\usepackage{xcolor}         
\usepackage{tikz}
\usetikzlibrary{automata,positioning,calc, arrows}
\usepackage{amssymb}
\usepackage{graphicx} 
\usepackage{natbib}
\setcitestyle{brackets}
\usepackage{booktabs}
\usepackage{amsmath}
\usepackage{bbm}
\usepackage{subcaption}
\usetikzlibrary{arrows.meta, positioning}
\title{PRIM: Meta-Learned Bayesian Root Cause Analysis}

\definecolor{iconcolor}{RGB}{140,160,185} 
\hypersetup{
    colorlinks=true,
    linkcolor=red,
    filecolor=magenta,      
    urlcolor=iconcolor,
    citecolor=blue,
}
\usepackage{fontawesome5}
\usepackage{xcolor}
\author{%
  Christopher Lohse$^{1,2}$\thanks{Correspondence to Christopher Lohse: \textcolor{iconcolor}\faEnvelope\ \href{mailto:lohsec@tcd.ie}{lohsec@tcd.ie}.},
  Anish Dhir$^{3}$,
  Amadou Ba$^{2}$,\\
  \textbf{Bradley Eck}$^{2}$, 
  \textbf{Marco Ruffini}$^{1}$,
  \textbf{Jonas Wahl}$^{4,5}$ \\ [4pt]
  $^{1}$University of Dublin, Trinity College, Dublin, Ireland \\
  $^{2}$IBM \\
  $^{3}$Gatsby Computational Neuroscience Unit, University College London \\
  $^{4}$Deutsches Forschungszentrum für Künstliche Intelligenz (DFKI), Saarbrücken, Germany \\
  $^{5}$Department of Philosophy, University of Bergen, Norway \\[4pt]
 }
\usepackage{amsthm}
\theoremstyle{definition}
\newtheorem{definition}{Definition}
\theoremstyle{definition}
\newtheorem{assumption}{Assumption}
\newtheorem{theorem}{Theorem}

\numberwithin{definition}{section}
\numberwithin{assumption}{section}
\newtheorem{proofthm}{Proof}
\begin{document}

\maketitle

\begin{abstract}
Root Cause Analysis (RCA) aims to identify the main drivers of anomalies in complex multivariate systems. The task is challenging due to the need for structural causal knowledge for some methods, and the computational cost of inference at test time. This paper introduces PRIM (\textbf{P}rior-fitted \textbf{R}oot cause \textbf{I}dentification with \textbf{M}eta-learning), a causal meta-learning approach that frames RCA as a Bayesian inference task over a prior of causal models. By marginalising out structural uncertainty, PRIM detects changes in the data-generating mechanism between normal and anomalous periods, infers distributional differences without explicit statistical testing, and implicitly learns causal structure without model fitting at test time. Following the simulation-based meta-learning paradigm of prior-fitted networks (PFNs), PRIM uses a transformer neural process (TNP) that jointly attends over normal and anomalous samples and the structural dependencies of nodes, enabling fast zero-shot inference in systems with up to 100 variables. Across synthetic benchmarks and two realistic benchmark datasets, PetShop and CausRCA, PRIM is competitive with other graph-free RCA methods.
\end{abstract}

\section{Introduction}
In complex multivariate dynamic systems, anomalies rarely occur alone. When a fault occurs, its effects propagate through the system, causing multiple variables to behave abnormally simultaneously. Root cause analysis (RCA) is the task of identifying the original source of an error based on observations of the system both before and during the anomalous period \citep{budhathoki2021did}. RCA is a critical task across many domains, including cloud microservices \citep{pmlr-v236-hardt24a, 10.1145/3308558.3313653, Pham2024}, manufacturing \citep{Mehling2026}, medicine \citep{selvaraj2019impact, lynn2011patterns} as well as climate and sustainability \citep{jayswal2011sustainability, belausteguigoitia2004causal}.
Existing approaches solve the RCA task either by assuming knowledge of a causal graph describing the dynamics of the system \citep{Li2022, Budhathoki2022, page1999pagerank} or by learning (parts of) a graph or statistical dependencies at inference time for example through causal discovery \citep{ikram2022root}.
Graph-based methods require accurate prior knowledge sometimes not available in practice, while methods that infer the graph at test time face computational costs scaling superexponentially with the number of variables \citep{pham2025rcaeval}. 
Graph-free approaches based on statistical testing such as $\varepsilon$-diagnosis \citep{10.1145/3308558.3313653} and RCD \citep{ikram2022root} come with the cost of a slower inference time, which scales superlinearly with the number of variables.




\begin{figure}
    \centering
    \includegraphics[width = \linewidth]{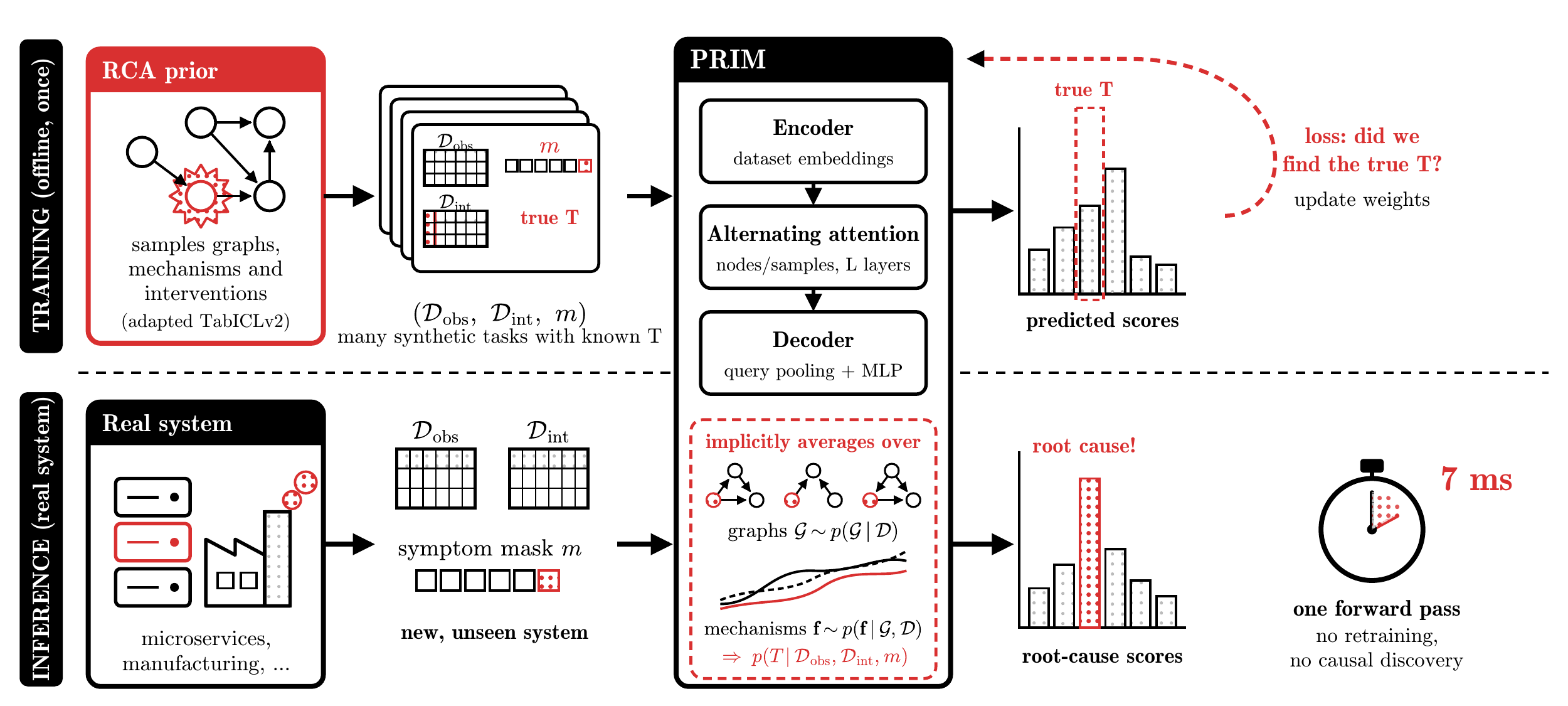}
    \caption{Simplified overview of the PRIM PFN training and inference process. The RCA prior samples synthetic data $(\mathcal{D}_{\mathrm{obs}}, \mathcal{D}_{\mathrm{int}}, m)$ with a known root cause $T$, and PRIM is trained until convergence. For a new real system, PRIM takes the observed data and symptom mask $m$ and returns root-cause scores in a single forward pass, implicitly averaging over the unknown graph $\mathcal{G}$ and mechanisms $\mathbf{f}$.}
    \label{fig:primov}
\end{figure}

A Bayesian, TNP (Transformer Neural Process)-based approach addresses these issues directly. Bayesian model-averaging over graph and functional mechanism uncertainty mitigates the finite-sample and identifiability issues that affect point-estimate causal discovery \citep{dhir2024meta, lorch2021dibs, cundy2021bcd, robertson2026dopfn} and attends to functional uncertainty over multiple models. The TNP collapses the usual two-step causal-discovery-then-RCA pipeline into a single inference step, removing the need for explicit distribution-shift testing \citep{10.1145/3308558.3313653, ikram2022root, pmlr-v236-hardt24a} and thus speeding up inference.
This motivates our re-framing of RCA as a Bayesian causal meta-learning problem \citep{dhir2024meta, Dhir2507, robertson2026dopfn}, following the simulation-based paradigm of prior-fitted networks (PFNs) \citep{hollmanntabpfn, muller2022transformers, qu2026tabiclv} which has proven effective for classification, anomaly detection \citep{Hollmann2025}, and time series forecasting \citep{hoo2025tables} with good results on real data \citep{Hollmann2025, qu2026tabiclv}. \autoref{fig:primov} shows a simplified overview of the Bayesian TNP for RCA.


\paragraph{Contributions} Our contributions are fourfold:

\begin{itemize}
    \item[1.] Inspired by \cite{Dhir2507, hollmanntabpfn, robertson2026dopfn}, we design and implement PRIM (\textbf{P}rior-fitted \textbf{R}oot cause \textbf{I}dentification with \textbf{M}eta-learning), a zero-shot inference method that ranks root causes given normal and anomalous data and does not require structural knowledge of the data-generating system as an input.
    \item[2.] To train PRIM, we adapt the TabICLv2 prior \citep{qu2026tabiclv} to the root cause analysis task. To generate anomalies, we complement TabICLv2 with a variety of structural interventions that are selected according to a two-sample test to ensure appropriate intervention strength.
    \item[3.] We discuss theoretical identifiability of root causes, adapting results by \cite{squires2020permutation, Jaber2020} to our setting.
    \item [4.] We empirically validate that PRIM converges to the true root cause given sufficient interventional/anomaly data, and evaluate PRIM on several synthetic setups and two realistic benchmarks, PetShop \citep{pmlr-v236-hardt24a} and CausRCA \citep{Mehling2026}. PRIM exhibits robust general performance across well-supported settings.
\end{itemize}


\section{Preliminaries}

\paragraph{Structural Causal Models}

A Structural Causal Model (SCM) quantitatively represents causal relationships among multiple random variables and describes how changes in a system variable \(X\) lead to changes in other variables \citep{pearl2009causality}. In an SCM, each variable \(X\) is assigned a value \(X := \phi_X(\mathrm{pa}_X, \epsilon_X)\) through a function $f$ of its \emph{causal parents} {\(\mathrm{pa}_X\)} and an independent noise term \(\epsilon_X\) \citep{pearl1995causal}. Qualitative causal relationships in an SCM can be represented graphically as a Directed Acyclic Graph (DAG) \(\mathcal{G}\) with directed edges $Z \to X, \ Z \in \mathrm{pa}_X$. \emph{Causal discovery} is the task to infer the structure of \(\mathcal{G}\) from observational or interventional data \citep{spirtes2016causal}.

\paragraph{Root Cause Analysis}
Consider variables $X = (X_1, \ldots, X_k)$ whose causal interactions are given by an SCM with underlying DAG  $\mathcal{G} = (V, E)$.
Following \citet{budhathoki2021did, ikram2022root} we define a \textbf{root cause} as a soft intervention on a subset of variables indexed by $T \subseteq \{1, \ldots, n\}$, resulting in a shifted distribution $P_X^T = \prod_{j \in T} \tilde{P}_{X_j \mid \text{pa}_j} \prod_{j \notin T} P_{X_j \mid \text{pa}_j}$, where $\tilde{P}_{X_j \mid \text{pa}_j}$ represents the modified conditional distribution for intervened variables.
Following \citet{eberhardt2007interventions}, we distinguish between intervention types: a \textit{hard intervention} $do(X_i = x)$ replaces the mechanism $f_i$ with a constant, eliminating the influence of $\text{pa}(X_i)$. In contrast, a \textit{soft intervention} replaces $f_i$ with a modified mechanism $f'_i$, altering the conditional distribution $P(X_i \mid \text{pa}(X_i))$ without removing the causal edges from its parents.
The \textit{root cause analysis task} is to recover the intervention target set $T$ given observational samples $\mathcal{D}^{\text{obs}} = \{x^{(k)}\}_{k=1}^{n_{\text{obs}}} \sim P_X$ (normal operating conditions), interventional samples $\mathcal{D}^{\text{int}} = \{x^{(k)}\}_{k=1}^{n_{\text{int}}} \sim P_X^T$ (anomalous conditions under intervention), and one or more observed symptom(s) $m \in \{0,1\}^K$ with $m_j = 1$  if the node $j$ is anomalous.
In our setup of RCA we consider that an observed anomaly/anomalies $m$ is known, which restricts the set of valid root causes $T$. Typically $m$ only contains one anomalous node, the 'problem' or 'target' node \citep{pmlr-v236-hardt24a}, but it some case it can also be several 'alarm nodes'\citep{Mehling2026}.

Throughout this paper we assume that we have access to $m$, which is the case in several RCA benchmark settings \citep{pmlr-v236-hardt24a, Mehling2026}.
Each sample $x^{(k)} = (x_1^{(k)}, \ldots, x_n^{(k)})$ is a realization of the variables $X = (X_1, \ldots, X_n)$. In applied root cause analysis settings, the observational data corresponds to normal system behaviour, while the interventional data represents anomalous observations collected when the system is malfunctioning or degraded \citep{ikram2022root, assaad2023root}.

\section{Bayesian Root Cause Analysis}
\citet{dhir2024meta, Dhir2507, robertson2026dopfn} formalise causal discovery and inference as Bayesian meta-learning problems, training a neural process to approximate posteriors over causal structures from data. We adapt their formulation to RCA: rather than firstly inferring $\mathcal{G}$, our goal is to infer the root cause set $T$ directly, avoiding a two-step pipeline of causal discovery followed by RCA.
An intervention on $T$ may induce a shift in functional relationships, noise distributions, or both, formalized as a change from the pre-intervention model $\mathbf{f}^{\text{obs}} = (\phi^{\text{obs}}, \epsilon^{\text{obs}})$ to the post-intervention model $\mathbf{f}^{\text{int}} = (\phi^{\text{int}}, \epsilon^{\text{int}})$. Here $\phi$ represents the collection of functional mappings and $\epsilon$ the collection of noise terms of the SCM.
The change is localized to the intervention target: $\mathbf{f}_i^{\text{int}} = \mathbf{f}_i^{\text{obs}}$ for all $i \notin T$, while $\mathbf{f}_i^{\text{int}} \neq \mathbf{f}_i^{\text{obs}}$ for $i \in T$. A Bayesian formulation of RCA avoids  committing to a single estimated DAG, propagating structural uncertainty about the DAG into the root cause posterior instead.

\paragraph{Bayesian Causal Model}
\label{def:bcm}
Adapting \citet{dhir2024meta, Dhir2507} to the RCA setting, we define a \emph{Bayesian causal model} (BCM) 
by drawing a DAG $\mathcal{G}$ from  $\mathcal{G} \sim p_{\text{BCM}}(\mathcal{G})$.
Conditional on $\mathcal{G}$, we sample the observational mechanism $\mathbf{f}^{\text{obs}} \sim p_{\text{BCM}}(\mathbf{f}^{\text{obs}} \mid \mathcal{G})$ and the root cause set $T \sim p_{\text{BCM}}(T \mid \mathcal{G})$.
The interventional mechanism is then drawn as $\mathbf{f}^{\text{int}} \sim p_{\text{BCM}}(\mathbf{f}^{\text{int}} \mid T, \mathbf{f}^{\text{obs}}, \mathcal{G}) = \prod_{i \in T} p_{\text{BCM}}(\mathbf{f}^{\text{int}}_i|\mathcal{G}) \prod_{j \notin T} \mathbbm{1}_{\mathbf{f}^{\text{int}}_j = \mathbf{f}^{\text{obs}}_j} $, reflecting that interventions modify only mechanisms or noises in $T$ and leave others invariant. The random variables $X^{\text{obs}}, X^{\text{int}}$ are distributed as $X^{\text{obs}} \sim p_{\text{BCM}}(X^{\text{obs}} \mid \mathbf{f}^{\text{obs}})$, respectively $X^{\text{int}} \sim p_{\text{BCM}}(X^{\text{int}} \mid \mathbf{f}^{\text{int}})$, and the corresponding datasets are generated as i.i.d.\ samples
\begin{equation*}
    \mathcal{D}^{\text{obs}} = \{ {X^{\text{obs}}}^{(k)} \}_{k=1}^{n_{obs}}, \ {X^{\text{obs}}}^{(k)} \sim X^{\text{obs}}, \qquad
    \mathcal{D}^{\text{int}} = \{ {X^{\text{int}}}^{(k)} \}_{k=1}^{n_{int}}, \ {X^{\text{int}}}^{(k)} \sim X^{\text{int}}.
\end{equation*}

We assume that $\mathrm{desc}(T) \subseteq \mathrm{anom}(X^\text{obs}, X^\text{int}) := \{ \ell \in V \mid p_{\text{BCM}}(X_{\ell}^{\text{int}} ) \neq p_{\text{BCM}}(X_{\ell}^\text{obs} ) \}$, meaning that each descendant of a root cause experiences a marginal distribution change\footnote{Observe that the converse inclusion $\mathrm{desc}(T) \supseteq \mathrm{anom}(X^\text{obs}, X^\text{int})$ holds by construction of our BCM.}. This can be considered a faithfulness assumption for root cause analysis as it precludes that the effect of the intervention cancels out along different paths to a descendant, 'coincidentally' reproducing the observational distribution. This assumption is implied by the $\mathcal{I}$-faithfulness assumption of \citep{squires2020permutation}, which is required for identifiability, see also Appendix \ref{app:id_proof}. As a consequence, this assumption allows us to sample the anomaly indicator $m$ uniformly from $\mathrm{desc}(T)$,  $m \sim p(m \mid \mathrm{desc}(T))$.

\paragraph{Bayesian Posterior and Identifiability of Root Causes}
Defining $\mathbf{f} \equiv (\mathbf{f}^{\text{obs}}, \mathbf{f}^{\text{int}})$, the Bayesian posterior over the root causes depends solely on the mechanism change, making separate component specifications redundant. The factorization of the distribution of the Bayesian Model gives the following conditional independence statements: 1) $T \perp X^{\text{obs}}, X^{\text{int}} \mid \mathbf{f}, m, \mathcal{G}$: given the mechanism change, graph, and anomaly location, the data carry no additional information about $T$. (2) $\mathcal{G} \perp m \mid X^{\text{obs}}, X^{\text{int}}$, and (3) $\mathbf{f} \perp m \mid X^{\text{obs}}, X^{\text{int}}$: the anomaly indicator carries no additional information about the graph or mechanism once $X^{\text{obs}}, X^{\text{int}}$ are known.
The Bayesian posterior over the root cause decomposes as
\begin{equation}
    p(T \mid \mathcal{D}^{\text{obs}}, \mathcal{D}^{\text{int}}, m) = \sum_{\mathcal{G}} \int p(T \mid \mathbf{f}, \mathcal{G}, m) \, p(\mathbf{f} \mid \mathcal{G}, \mathcal{D}^{\text{obs}}, \mathcal{D}^{\text{int}}) \, p(\mathcal{G} \mid \mathcal{D}^{\text{obs}}, \mathcal{D}^{\text{int}}) \, d\mathbf{f}, 
\end{equation}
which marginalises over graph and functional mechanism uncertainty. See Appendix~\ref{app:post}) for the computation of this expression. 

\begin{theorem}[Identifiability of Root Causes]\label{thm:identifiability}
Assume that the data-generating SCM is causally sufficient with acyclic underlying DAG, and that mechanisms remain invariant for non-intervened nodes as in the definition of a BCM. Under $\mathcal{I}$-faithfulness (see Appendix \ref{app:id_proof}), $p_{\text{BCM}}(X_{\ell}^\text{obs})$ and $p_{\text{BCM}}(X_{\ell}^\text{int})$ together
with the anomaly indicator $m$ identify the root-cause set $T$, even though the underlying DAG itself may
not be uniquely identifiable.
\end{theorem}
This result is a direct adaptation of the results of \cite{squires2020permutation} (see also \cite{Jaber2020, ikram2022root} for parallel results) to our setting. A full proof for \autoref{thm:identifiability} is provided in Appendix~\ref{app:id_proof}.

Bayesian model averaging is well-suited to this setting: it correctly propagates the residual graph ambiguity into the posterior over $T$.
The posterior is intractable as the sum over $\mathcal{G}$ is exponential in $|V|$ and the integral over $\mathbf{f}$ is over an infinite-dimensional function space.
Following \citet{Dhir2507,dhir2024meta, robertson2026dopfn}, we therefore amortise inference with a prior fitted transformer neural process model $f_\theta$, trained on synthetic $BCM$ samples, that directly scores candidate root cause nodes
\begin{equation}
p_{\theta}(T = j \mid \mathcal{D}^{\text{obs}}, \mathcal{D}^{\text{int}}, m) = \text{softmax}\!\left(f_{\theta}(\mathcal{D}^{\text{obs}}, \mathcal{D}^{\text{int}}, m)\right)_{j}, \quad j \in V.
\end{equation}

Under standard assumptions on the mass concentration of the prior, posterior updates  concentrate asymptotically on the true root cause set as the sample size goes to infinity \citep{ghosal_vandervaart_2017_consistency}. The results of \cite{hollmanntabpfn, qu2026tabiclv, dhir2025continuous} suggest empirically that TNPs are good approximators of Bayesian posterior distributions.




\section{Related Work}

\paragraph{Transformer Neural Process and Prior-Fitted Networks}
Since estimating the posterior over graph and functional uncertainty is hard, we use a Transformer Neural Process (TNP) trained as a Prior-Fitted Network (PFN), an approach proven effective on similar problems \cite{hollmanntabpfn, Hollmann2025, dhir2025continuous, dhir2024meta, robertson2026dopfn} across diverse domains \cite{Nguyen2022, feng2023latent, Ashman2024}. PFN refers to the training paradigm: learning to approximate the posterior predictive by fitting to samples drawn from a synthetic prior. TNP refers to the architecture, a Neural Process built on the transformer \citep{Vaswani2017}. Trained this way, PFNs learn general tabular representations with strong performance on real world benchmarks, even though its trained on synthetic data on regression and classification \cite{hollmanntabpfn, hoo2025tables, muller2022transformers}, can be fine-tuned to new distributions \cite{hoo2025tables}, and as Neural Processes produce uncertainty-aware estimates at inference without intermediate steps such as explicit graph estimation \citep{Dhir2507}. They have been applied to causal discovery \cite{dhir2025continuous} and effect estimation \cite{Dhir2507, robertson2026dopfn}; in particular, \citet{Dhir2507, robertson2026dopfn} use a TNP to estimate interventional distributions given an intervention target, making it a natural foundation for estimating $T$ directly from $D^{\text{obs}}$, $D^{\text{int}}$, and $m$ without a separate causal discovery step.
Using a Bayesian TNP avoids two-step algorithms that first fix a graph and then perform RCA, instead inferring $T$ jointly with $\mathcal{G}$ and the mechanisms.
Further, it does not rely on statistical distribution difference tests to detect mechanism changes, which are often underpowered under limited interventional data.

\paragraph{Root Cause Analysis}
RCA closely aligns with intervention target estimation~\citep{Jaber2020, varici2021scalable, varici2022latent, yang2024learning, squires2020permutation}: in both, interventions on a small set of nodes explain shifts in the overall distribution, and the goal is to identify these nodes. It has been applied to cloud microservices \citep{soldani2022anomaly, chen2014causeinfer, lin2018microscope, 10.1145/3308558.3313653, liu2021microhecl, Li2022, Pham2024, Liang2026, pmlr-v236-hardt24a}, manufacturing \citep{Mehling2026}, medicine \citep{selvaraj2019impact, lynn2011patterns}, and sustainability \citep{jayswal2011sustainability, belausteguigoitia2004causal}.
Existing approaches differ in whether they assume a known causal graph, and most impose some parametric assumption on the mechanism family.

\emph{Graph-based} methods require the true causal graph as input. Traversal \citep{pmlr-v236-hardt24a} selects the highest-ranked anomalous node with no anomalous parent connected to $m$. CIRCA \citep{Li2022} fits a linear Gaussian SCM and tests which conditional distribution has changed. \citet{Budhathoki2022} attribute root causes counterfactually via Shapley values; \citet{nagalapatti2025robust} later replace the counterfactual estimates with interventional ones for improved robustness to out-of-distribution anomalies. PageRank \citep{page1999pagerank} and CausTR \citep{Mehling2026} instead rank variables by graph proximity: PageRank by importance in the graph, and CausTR by causal proximity to alarm activations.

\emph{Graph-free} methods relax this requirement, but typically introduce other assumptions. RCD \citep{ikram2022root} performs local causal discovery under unknown soft interventions \citep{Jaber2020, squires2020permutation}, later extended to settings with partial structural knowledge \citep{ikram2025partial}. RUN \citep{lin2024root} and \citet{han2025root} instead recover a causal graph via neural Granger causal discovery from time series, combined with PageRank-based and integrated Granger-causal RCA, respectively. The remaining methods are graph-free by design: $\varepsilon$-diagnosis \citep{10.1145/3308558.3313653} uses two-sample tests, BARO \citep{Pham2024} ranks by deviation magnitude, TimeRank \citep{Mehling2026} ranks by recency of value changes, and Meta-RCA \citep{Liang2026} combines LLMs with a Bayesian belief model and an LLM-derived knowledge graph.


PRIM differs from all the above method categories. In difference to graph-based methods such as Traversal, CIRCA, PageRank, CausTR and Shapley attribution, it requires no causal graph. In contrast to graph-free methods that estimate a graph at test time (RCD and RUN), its cost does not increase with the search over structures. Unlike methods that rely on explicit distribution-shift testing (CIRCA, $\epsilon$-diagnosis) or deviation heuristics (BARO, TimeRank, CausTR), PRIM's ranking is a single amortised posterior estimate. This retains a probabilistic treatment of graph and mechanism uncertainty via Bayesian model averaging, rather than committing to a point estimate at any stage.

\section{Prior and Model Architecture}
The following section outlines the design of our main contributions: adapting a prior of SCMs for the RCA setting and defining a transformer neural process architecture that learns to approximate the Bayesian posterior of root causes.

 \subsection{Adapting a prior for the RCA setting}\label{sec:setup}

Specifying a good prior is crucial for any PFN to work in real-life settings using the simulation-to-real approach \citep{hollmanntabpfn, qu2026tabiclv}. Unlike other priors, which are only used to do classification, regression, or causal inference \citep{hollmanntabpfn, qu2026tabiclv, dhir2024meta}; for root cause analysis the distribution shift in the anomalous data needs to resemble real-life failures in complex systems. We propose that a good prior for RCA must satisfy the following three criteria: 
\begin{enumerate}
    \item The underlying BCM must sample from a graph whose density and structure resemble the system to be estimated.
    \item The mechanisms for the structural equations must resemble the underlying mechanisms observed during inference.
    \item The intervention specification, i.e. the shift causing the shift in the root cause variable, must resemble the shifts we expect to see during inference.
\end{enumerate}

\begin{figure}[h!]
    \centering
    \includegraphics[width=1\linewidth]{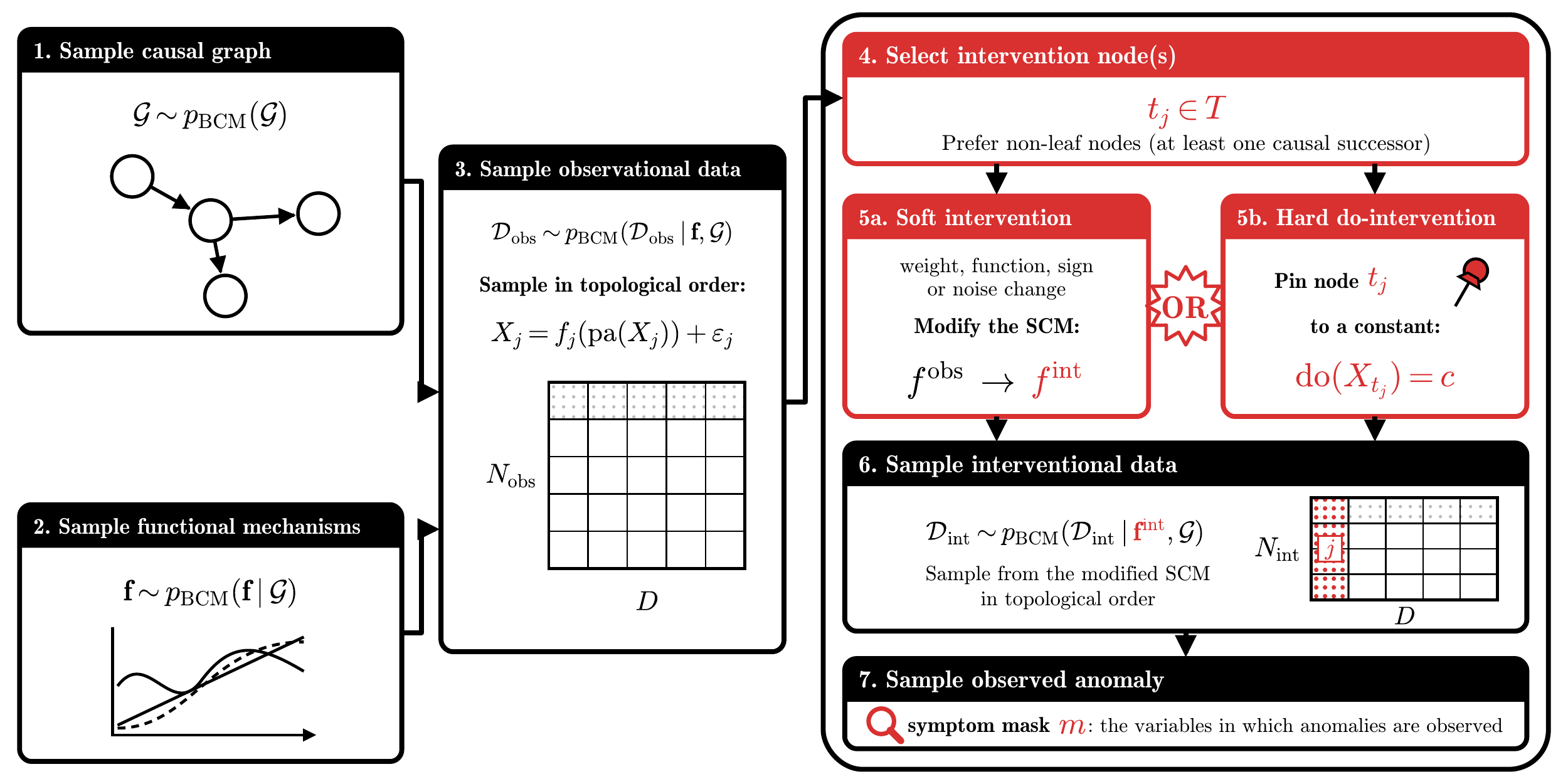}
    \caption{Overview of the data generation process for the prior. A causal graph $\mathcal{G}$
    and functional mechanisms $\mathbf{f}$ are sampled to define an SCM.
    Observational data $\mathcal{D}_{\text{obs}}$ is drawn by ancestral sampling,
    after which a target node $t_j \in T$ is selected and the SCM is perturbed
    via a soft intervention ($\mathbf{f}^{\text{obs}} \!\to \mathbf{f}^{\text{int}}$)
    or a hard do-intervention pinning $t_j$ to a constant.
    Interventional data $\mathcal{D}_{\text{int}}$ is then sampled from the
    modified SCM. The final sample is $(D^{\text{obs}}, D^{\text{int}}, m)$.}
    \label{fig:datagen_prior}
\end{figure}

The steps of the data sampling process for generating data from the  prior is shown in \autoref{fig:datagen_prior}. 
To generate the causal Graph that underlies the data-generation process, we sample sparse DAGs (Erd\H{o}s--R\'enyi, Barab\'asi--Albert, or bipartite) over $K\!\sim\!\mathrm{Unif}(K_{\min},K_{\max})$ nodes with expected in-degree $\bar{d}\!\sim\!\mathrm{Unif}(1.8,2.5)$. We do so to satisfy the first point for a good RCA prior: as the systems usually observed for RCA are sparse, the in-degree is based on common values for real systems. 

As a starting point for generating the observational distribution to satisfy point two, we use the TabICLv2 Prior \citep{qu2026tabiclv}, since this prior design has shown promising performance on real data in earlier work \citep{qu2026tabiclv}, and offers a broad set of different distributions. We keep the functional mechanisms (mixed MLP, tree, linear, and Gaussian process) of TabICLv2, and also use its discrete data sampling. We further add temporal structure to the prior, with lower noise and higher autocorrelation than the other used distributions, as many RCA datasets are time series.

As our method takes as input the pre-anomaly data $\mathcal{D}^{\text{obs}}$ as well as the anomalous data $\mathcal{D}^{\text{int}}$, we need to specify the types and effects of interventions included in the prior. This is not included in the TabICLv2 prior, since it never samples interventional data. This is one of the contributions made by us.
Following standard synthetic RCA data generation \citep{assaad2023root}, we differentiate between interventions that shift the weight of the structural equations, shift the mean, inflate the noise. We further add a fixed value do-intervention and sample 1 to $t_{max}$ (maximum number of parallel faults) root causes at a time. 
 For higher realism, we also introduce sampling failures that are restricted to a local neighbourhood and do not affect the majority of the system.

To select an appropriate intervention strength, we run a Maximum Mean Discrepancy (MMD) two-sample permutation test \citep{JMLR:v13:gretton12a} across a grid of different intervention strengths. We select the smallest magnitude where the mean test rejection rate exceeds $0.9$ at a significance level of $\alpha = 0.05$, ensuring that the induced distributional shift is detectable and learnable by any estimator. 
We randomly sample these types of interventions in each training pair generation, following our setup of the BCM defined in \autoref{def:bcm}. In addition, we sample multiple causally independent root causes.  We assume that the root cause $T$ is never a leaf node nor the observed anomaly $m$ itself. Each training samples samples $(\mathcal{D}^{obs}, \mathcal{D}^{int}, m)$ from the prior $p(\mathcal{M})$. Our prior satisfy our identifiability assumptions; for an in-depth description of the prior design, the data sampling process, and all hyperparameters, see Appendix \ref{app:data-gen}.


\subsection{Model Architecture}

\begin{figure}[h!]
    \centering
    \includegraphics[width=1\linewidth]{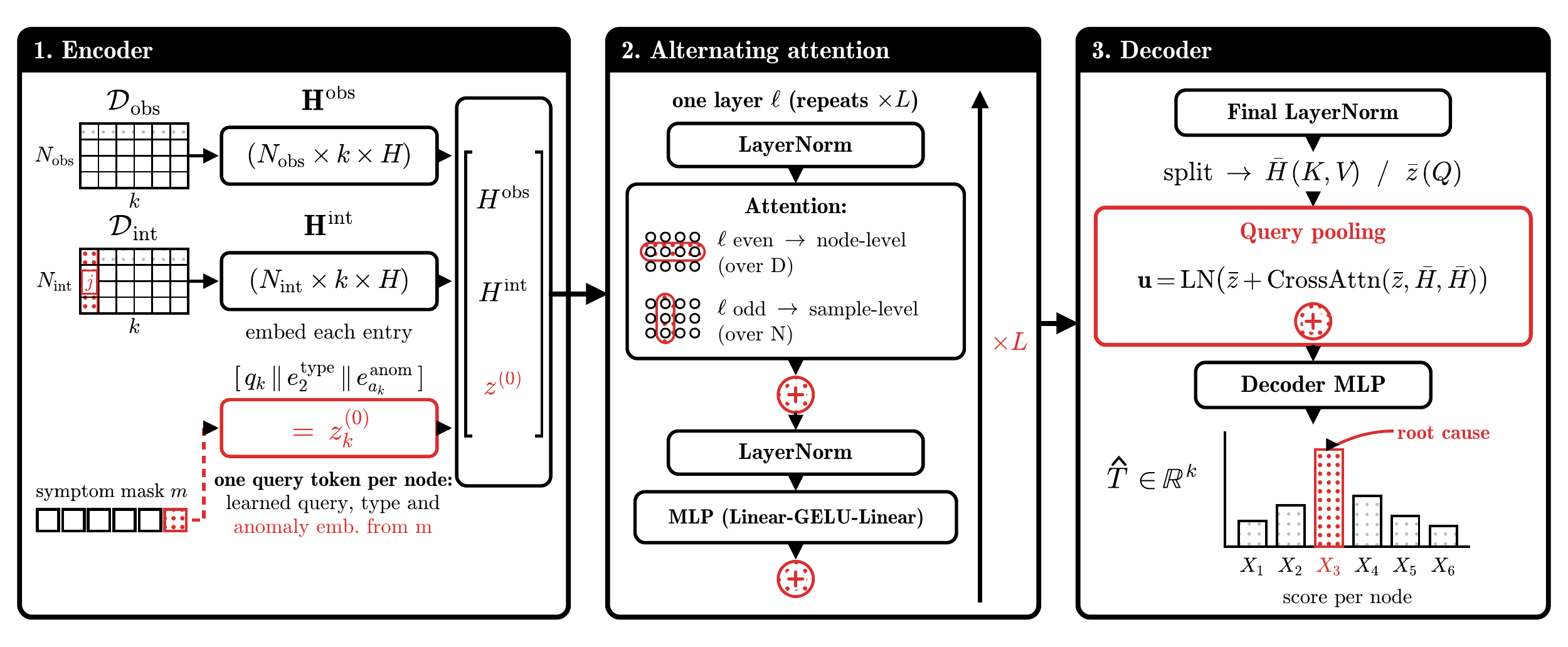}
    \caption{PRIM-NeuralProcess architecture (see text for description). $K$: number of nodes; $N$: number of samples in the sequence (observational + interventional + the query token); $z^{(0)}_k = [q_k \,\Vert\, e^{\text{type}}_2 \,\Vert\, e^{\text{anom}}_{a_k}]$: learned per-node query token, with $a_k = m_k$ if the optional symptom mask $m$ is given, else a fixed ``not provided'' index; ``+'': residual (skip) connection; $\bar{z}, \bar{\mathbf{H}}$: query/samples after the final layer norm; $\mathbf{u}$: per-node summary produced by QueryPooling; $\hat{T} \in \mathbb{R}^k$: output logits.}
    \label{fig:arch}
\end{figure}

Our model, PRIM (\textbf{P}rior-fitted \textbf{R}oot cause \textbf{I}dentification with \textbf{M}eta-learning), predicts which node of a system was intervened on (the root cause) from paired observational and interventional data, using an architecture drawing inspiration from the causal-structure-induction model of \citet{ke2023learning}. An overview of the architecture is given in \autoref{fig:arch}. 
Observational and interventional samples are each encoded into per-node embeddings by concatenating a data-type-specific value projection, a fixed sinusoidal node-identity code, a learned sample-type tag (observational/interventional), and a learned anomaly tag derived from an optional binary symptom mask $m$. A trainable per-node query token $z^{(0)}$ is appended to the resulting sequence, and $L$ pre-norm transformer blocks refine it via attention that alternates between attending across the $K$ nodes and attending across the $N$ samples (observational, interventional, and the query). This enables observational and interventional samples attend to each other bidirectionally at every layer. After the final layer norm, an attention layer reads out the evolved query $\bar{z}$ over the evolved samples $\bar{\mathbf{H}}$ to produce a per-node summary $\mathbf{u}$. A two-layer MLP decodes $\mathbf{u}$ into per-node logits $\hat{T}\in\mathbb{R}^k$.

The model is trained on synthetic SCMs drawn from $p(\mathcal{M})$ to predict the (possibly multiple) true root-cause nodes $T \subseteq \{1,\dots,k\}$, minimizing the cross-entropy
\begin{equation}
    \mathcal{L} = \mathbb{E}_{\mathcal{M}}\, \mathbb{E}_{T, \mathcal{D} \mid \mathcal{M}} \Bigg[-\sum_{i=1}^{k} y_i \log \hat{y}_i\Bigg], \qquad y_i = \frac{\mathbf{1}[i \in T]}{|T|},
    \label{eq:loss}
\end{equation}
where $\hat{y} = \mathrm{softmax}(\hat{T})$ and $y$ is a uniform target distribution over the true root-cause node(s). This reduces to the standard one-hot cross-entropy when $|T|=1$. An overview is shown in Figure~\ref{fig:arch}; full architectural details are given in Appendix~\ref{app:architecture}.

\section{Experiments}
In order to evaluate PRIM we ran several experiments across simulated and realistic benchmarks. We first test for empirical convergence, then for different system sizes and finally for realistic data with unknown distributions. We use two different models for the experiment: PRIM-S a smaller model is used for experiments with up to $k_{max}=10$ and PRIM-L is used for the other experiments with up to $k_{max}=100$ variables the detailed parameters are given in Appendix \ref{app:hyperparams}.
\subsection{Synthetic Data}
In this section, we evaluate the performance of PRIM on RCA problems that resemble its training set, using GP and neural network mechanisms to cover a wide range of nonlinear functions. We report 90\% bootstrap confidence intervals (based on 500 resamples) over 200 episodes.

\begin{figure}[h!]
    \centering
    \begin{subfigure}{0.10\textwidth}
        \centering
        \begin{tikzpicture}[
            node distance=0.8cm,,
            main/.style  = {draw, shape=circle, minimum size=0.55cm, inner sep=0pt, font=\small},
            root/.style  = {draw, shape=circle, minimum size=0.55cm, inner sep=0pt, font=\small, fill=red!70},
            desc/.style  = {draw, shape=circle, minimum size=0.55cm, inner sep=0pt, font=\small, fill={rgb,255:red,255; green,179; blue,102}}
        ]
            \node[desc] (Zm) {Z};
            \node[root] (Xm) [below left=0.7cm and 0.2cm of Zm] {X};
            \node[desc] (Ym) [below right=0.7cm and 0.2cm of Zm] {Y};
            \draw[->] (Xm) -- (Zm);
            \draw[->] (Xm) -- (Ym);
            \draw[->] (Zm) -- (Ym);
            \node[draw=none, font=\scriptsize\normalfont, above=0.1cm of Zm] {Mediator};
        \end{tikzpicture}
        \begin{tikzpicture}[
            node distance=0.8cm,
            main/.style  = {draw, shape=circle, minimum size=0.55cm, inner sep=0pt, font=\small},
            root/.style  = {draw, shape=circle, minimum size=0.55cm, inner sep=0pt, font=\small, fill=red!70},
            desc/.style  = {draw, shape=circle, minimum size=0.55cm, inner sep=0pt, font=\small, fill={rgb,255:red,255; green,179; blue,102}}
        ]
            \node[main] (Zc) {Z};
            \node[root] (Xc) [below left=0.7cm and 0.2cm of Zc] {X};
            \node[desc] (Yc) [below right=0.7cm and 0.2cm of Zc] {Y};
            \draw[->] (Zc) -- (Xc);
            \draw[->] (Zc) -- (Yc);
            \draw[->] (Xc) -- (Yc);
            \node[draw=none, font=\scriptsize\normalfont, above=0.1cm of Zc] {Confounder};
        \end{tikzpicture}
    \end{subfigure}%
    \hspace{0.02\textwidth}
    \begin{subfigure}{0.38\textwidth}
        \centering
        \includegraphics[width=\linewidth]{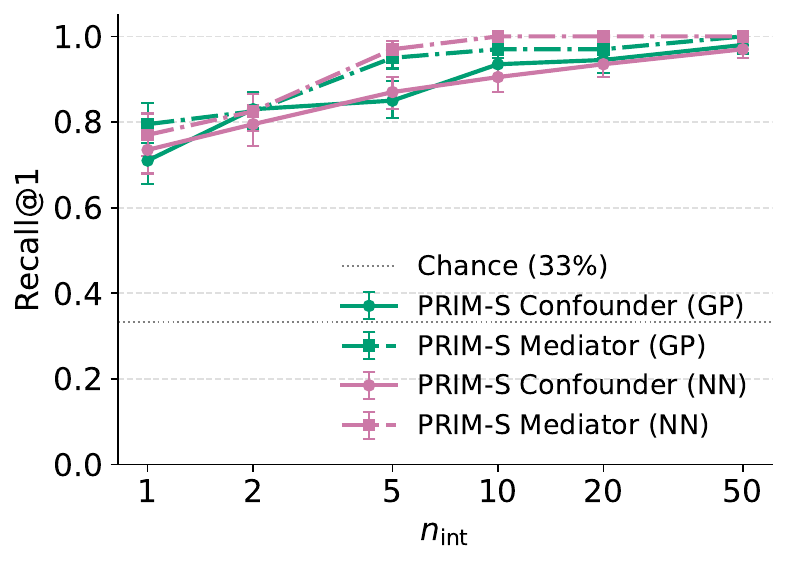}
        \caption{Varying $n_{\text{int}}$, keeping $n_{\text{obs}} = 100$}
        \label{fig:threenode_nint}
    \end{subfigure}
      \hspace{0.02\textwidth}
    \begin{subfigure}{0.38\textwidth}
        \centering
        \includegraphics[width=\linewidth]{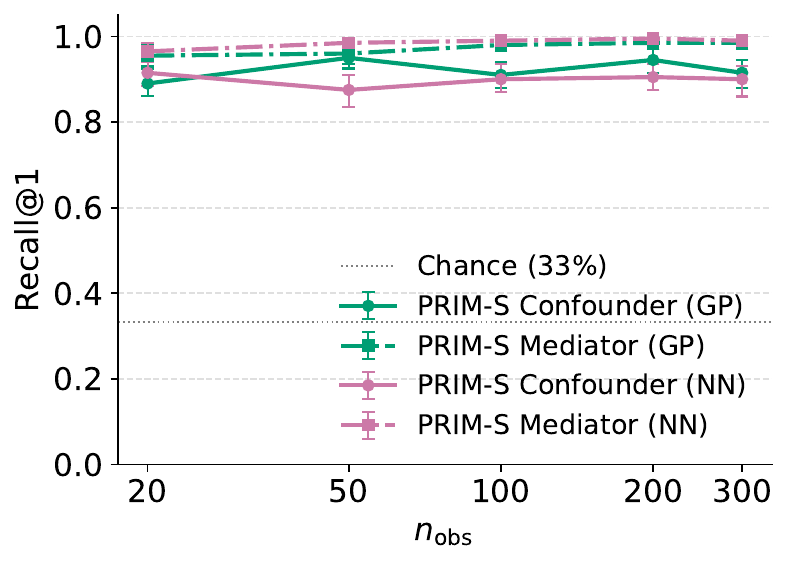}
        \caption{Varying $n_{\text{obs}}$, keeping $n_{\text{int}} = 10$}
        \label{fig:threenode_nobs}
    \end{subfigure}%

    \vspace{-0.5em}
    \caption{Three-node confounder vs.\ mediator scenario. X ({\color{red!70}$\blacksquare$} red) is the root cause; downstream nodes ({\color{rgb,255:red,255;green,179;blue,102}$\blacksquare$} orange) are descendants; unaffected nodes are white.}
    \label{fig:gp vs nn}
\end{figure}

\paragraph{Empirical Convergence Study}
In the first experiment, we evaluate in a controlled environment if PRIM correctly identifies the true root cause given sufficient interventional and observational data. This is an empirical sanity check of our analytical identifiability result.
As we want to test a standard RCA setup we always assume fewer anomalous samples $n_{int}$ than normal samples $n_{obs}$.
In \autoref{fig:threenode_nint}, we test if, for a fixed $n_{obs} = 100$ and a varying $n_{int}$, more anomalous samples lead to convergence of the model to the true root causes, and the same for a fixed $n_{int} = 10$ (a realistic number of anomalous samples) by varying $n_{obs}$.
We test on two three-node structures shown in \autoref{fig:gp vs nn}: a mediator and a confounder, with $X$ always representing the root cause.  This tests whether PRIM correctly identifies the root cause in the presence of indirect causal paths and common causes.
To generate the data, we use a Gaussian process and a neural network as structural functions. In the mediator setting (root cause is a root node), we intervene by inflating its noise, for both mechanisms. In the confounder setting (root cause has a parent), we instead scale the structural function generating it from its parent: its output scale for the Gaussian process, its weights for the Neural Network.

\autoref{fig:threenode_nint} shows that for both noise functions PRIM converges to the true root cause given enough interventional data. \autoref{fig:threenode_nobs} illustrates that more observational data does help the model to converge to the true root cause given a limited amount of $n_{int}$. We also tested this for a multi-RCA setting, and out-of-distribution generalisation with the same results: given enough interventional data, PRIM converges to the true root cause set. The results are provided in Appendix \ref{app:multirca} and \ref{app:finetune} respectively.

\begin{figure}[h!]
\centering
\begin{subfigure}[b]{0.4\textwidth}
    \centering
    \includegraphics[height=3.8cm]{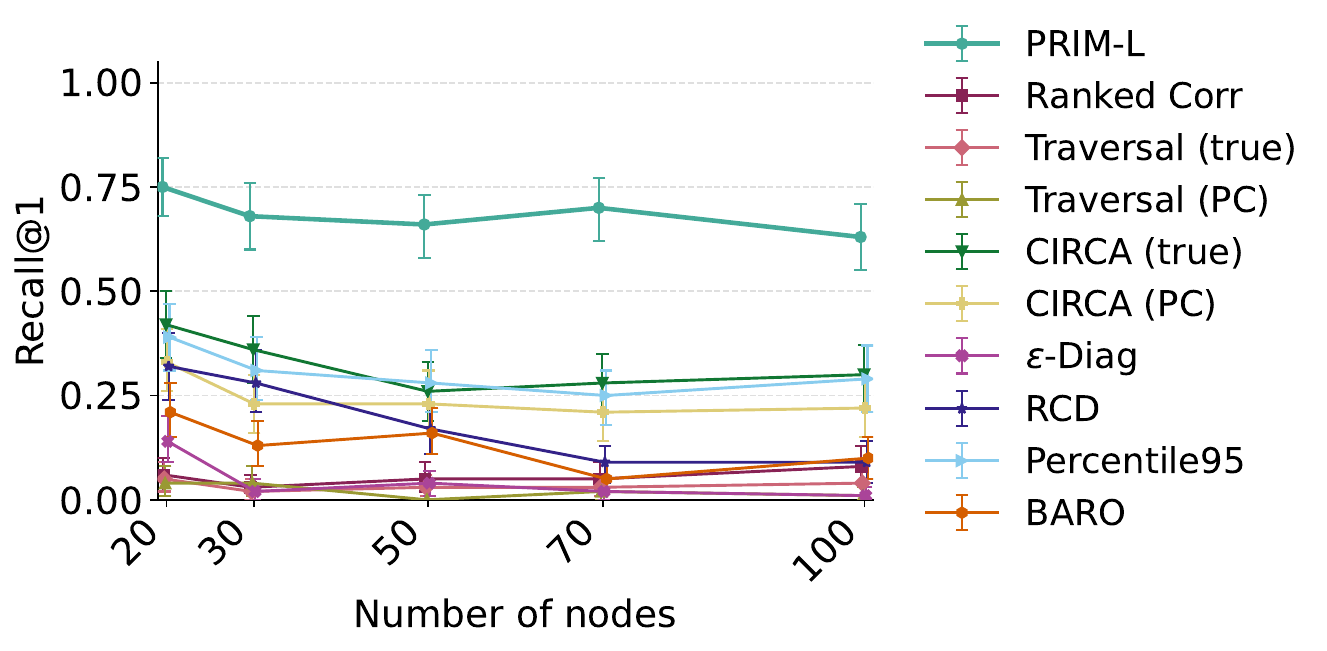}
    \caption{Gaussian Process Functions}
    \label{fig:errorbar_gp}
\end{subfigure}%
\hspace{0.15\textwidth}
\begin{subfigure}[b]{0.4\textwidth}
    \centering
    \includegraphics[height=3.8cm]{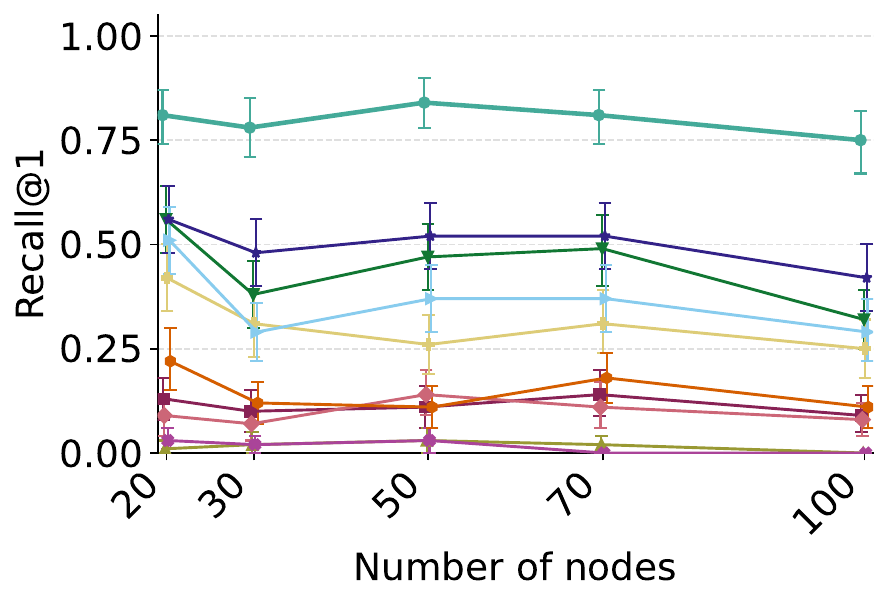}
    \caption{Neural Network Functions}
    \label{fig:errorbar_nn}
\end{subfigure}
\vspace{-0.5em}
\caption{Recall@1 vs.\ number of nodes at $n_{\mathrm{obs}}=100$, $n_{\mathrm{int}}=10$.}
\label{fig:errorbar_n100}
\end{figure}

\paragraph{Different System Sizes}To assess scalability, we evaluate PRIM across randomly sampled SCMs of varying size at $n_{\mathrm{obs}}=100$ and $n_{\mathrm{int}}=10$, with performance measured by Recall@1 (
\autoref{fig:errorbar_n100}). We include noise distributions that are identical to the prior PRIM was trained on to assess that its ability to detect in distribution, these distribution might not fit some parametric/linearity assumption of some of the methods compared against.
The intervention setup follows the calibrated prior described in Section~\ref{sec:setup}: Weight-change and additive shift rescale or offset the node's output after its structural equation is evaluated, identically for the GP and the NN; hard do-intervention replaces that output with a constant; and noise inflation scales the exogenous noise term added inside the mechanism itself, on top of the NN's forward pass or the GP's sampled kernel value respectively.

We also include PC-versions of graph-based methods (PC-Traversal, PC-CIRCA), where the causal graph is estimated via the PC algorithm \citep{spirtes2000causation}, providing graph-free variants of these methods for a fair comparison. Further, we add Percentile95 as a simple benchmark introduced by \cite{fang2026rethinking}, which ranks the root causes based on the maximum difference between anomalous samples and the 95 percentile of the distribution of the normal data.

As highlighted in  \autoref{fig:errorbar_n100}, PRIM outperforms all baselines in both GP and NN settings, including methods with access to the ground truth graph, and its Recall@1 degrades more slowly with increasing graph size than any other method. Recovering $T$ from the GP data is harder across the board due to weaker distributional shifts compared to NN weight-change interventions. CIRCA is the strongest competitor in the GP setting, while for NN data CIRCA leads for smaller graphs (up to ${\sim}20$ nodes) and RCD becomes competitive at larger sizes. The 95Percentile simple benchmark performs surprisingly well on both settings, similar to CIRCA, which has access to the true graph, in the GP setting. 
The PC-based variants are substantially worse than their True-graph counterparts, likely due to nonlinear functional mechanisms that violate the assumptions of the executed version of the PC algorithm,  even though they still outperform RCD in the GP and $\varepsilon$-diagnosis (which also assume linear relationships) in both settings.
The full results over different interventional sizes and the small model can be found in Appendix \ref{app:diff_nodes}.

Since inference time is crucial for real-world deployment, we benchmarked all methods across node sizes. PRIM stays constant at 7ms (large model) as it requires only a single forward pass, while other methods scale exponentially. Full results are in \ref{app:time}.
On a MacBook Pro M3 Max, PRIM runs in ${\sim}123$ms on CPU and ${\sim}17$ms on MPS GPU across all node sizes.

\subsection{Realistic Data with unknown Distribution}\label{sec:real}
We evaluate on two real-world benchmarks spanning cloud microservices and
manufacturing.
The PetShop benchmark \citep{pmlr-v236-hardt24a} (CC Public License)
is a microservice dataset based on the AWS Observability Workshop comprising
${\sim}44$ services monitored via latency and availability metrics, with 48
fault injection issues across four traffic scenarios, each with a single root
cause. $\mathcal{D}^{\text{obs}}$ covers the normal baseline period
($|\mathcal{D}^{\text{obs}}| \in \{589, 1652\}$ depending on scenario) and
$\mathcal{D}^{\text{int}}$ the anomaly period ($|\mathcal{D}^{\text{int}}| = 5$
five-minute snapshots); the symptom mask $m$ indicates the entry-point service
(\textsc{PetSite}).
The CausRCA benchmark \citep{Mehling2026} (Apache License 2.0) is a
partially real, partially simulated factory system with 100 runs across 19 fault
injection experiments in three subsystems (probe: 34, coolant: 25, hydraulics:
41), featuring multiple simultaneous root causes; all firing alarm nodes are
included in $m$. On average $|\mathcal{D}^{\text{obs}}| = 196$ and
$|\mathcal{D}^{\text{int}}| = 123$ (500\,ms discretisation).
Model weights are never updated on either benchmark; we use PRIM-L throughout. We only report the average values on these data sets here.

\begin{table}[h!]
\centering
\caption{Average accuracy of graph-not-given RCA methods on PetShop (top-3 recall, averaged over six traffic/metric scenarios) and CausRCA (MAP@3, averaged over three systems; \emph{Full} and \emph{Sub}graph settings shown separately). Results marked * are from the original papers~\cite{pmlr-v236-hardt24a,Mehling2026}; -- indicates not reported. \textbf{Bold} is the best result per row. \textsuperscript{\textdagger}~denotes our method.}
\label{tab:avg-results-combined}

\begin{tabular}{ll r@{.}l r@{.}l r@{.}l r@{.}l r@{.}l c r@{.}l}
\toprule
& & \multicolumn{2}{c}{$\varepsilon$-diag} & \multicolumn{2}{c}{RCD} & \multicolumn{2}{c}{TimeRank} & \multicolumn{2}{c}{Baro} & \multicolumn{2}{c}{Corr} & CausTR+PC & \multicolumn{2}{c}{PRIM\textsuperscript{\textdagger}} \\
\midrule
\multicolumn{2}{l}{PetShop} & 0&10\textsuperscript{*} & 0&39\textsuperscript{*} & 0&15 & 0&08 & 0&78\textsuperscript{*} & -- & \textbf{0}&\textbf{79} \\
\midrule
\multirow{2}{*}{CausRCA} & Full & 0&02 & 0&04 & 0&24 & 0&09 & 0&05 & -- & \textbf{0}&\textbf{27} \\
& Sub  & 0&11 & 0&24 & 0&60 & \textbf{0}&\textbf{75} & 0&33 & 0.57\textsuperscript{*} & 0&61 \\
\bottomrule
\end{tabular}%

\end{table}

As shown in Table~\ref{tab:avg-results-combined} among methods that do not require a known causal graph, PRIM achieves the highest average recall, outperforming all graph-not-given baselines.
Based on the reported MAP@3 in Table~\ref{tab:avg-results-combined}. PRIM is the best graph-not-given method on the full system and is slightly worse than Baro on the subsystem, giving the overall  most balanced performance. A full set of the detailed results, including a comparison with graph-based methods can be found in Appendices \ref{app:pet} and \ref{app:causrca}. On both datasets PRIM shows a balanced performance in line with other methods which highlights its general application.

\section{Conclusion}
In this paper we reformulate root cause analysis as a causal meta-learning problem and introduce PRIM, a model that identifies root causes via a single forward pass through a transformer neural process. PRIM approximates a Bayesian model-averaged posterior over graph and functional mechanism uncertainty using synthetic priors, effectively performing a parameter-free Bayesian test at inference time without any statistical testing or model fitting on real data.
We present a way to design a prior for RCA and a TNP architecture that is able to estimate the true posterior of root causes based on the learned prior. PRIM is competitive with RCA baselines across diverse datasets and domains without requiring a causal graph or retraining at deployment time. Our synthetic experiments empirically confirm that $\mathcal{D}^{\text{int}}$ and $m$ can be used to identify $T$: PRIM converges to the true root cause(s) given sufficient interventional data. On the realistic benchmarks PetShop~\cite{pmlr-v236-hardt24a} and CausRCA~\cite{Mehling2026}, PRIM is competitive with other graph-free RCA methods. Our paper highlights the usefulness of meta-learned PFNs for root cause analysis.

\paragraph{Limitations} Our approach has several limitations. First, identifiability of $T$ rests on the assumptions stated in Appendix~\ref{app:id_proof}: an acyclic SCM, causal sufficiency, mechanism invariance for non-targets (these are intrinsic to our modeling approach), $\mathcal{I}$-faithfulness, and the requirement that the observed anomaly is induced by the mechanism changes. Violations such as latent confounding or faithfulness violations compromise theoretical identifiability. Second, a misspecified prior can degrade accuracy by failing to capture the true RCA distribution as observed in the full causRCA example. 
Finally, PRIM cannot exploit potentially available (partial) knowledge of the ground-truth graph. 

\tikzset{>=latex}

\begin{ack}
We would like to thank the anonymous reviewers for their helpful feedback and a constructive discussion which we believe helped shping this work into a better paper. \\ 
C.L. was partilaly funded by the EcoTwin project, which has received funding from the European Commission's Horizon Europe programme under grant agreement No 101157369.\\
JW received support from the German Federal Ministry of Research, Technology, and Space (BMFTR) as part of the project MAC-MERLin (Grant Agreement No. 01IW24007) and from the European Regional Development Fund (ERDF) and the German Federal State of Saarland within the scope of (To)CERTAIN (Project ID EFRE-AuF-0000942).
\end{ack}

\bibliographystyle{plainnat}
\bibliography{bib.bib}


\appendix
\makeatother
\counterwithin{table}{section}
\counterwithin{figure}{section}
\renewcommand{\theequation}{\thesection\arabic{equation}}
\numberwithin{equation}{section}

\clearpage
\section{Changes in the Camera-Ready Version}
Based on the discussions and reviews on \href{https://openreview.net/forum?id=edcYZi5wbN#discussion}{openreview} we made the following changes in the camera ready version and reran all experiments:

\begin{enumerate}
    \item We improved the various unclear sections in the literature review and background sections.
    \item We adapted the construction of the prior and clarified every step in the selection process of the prior, as requested by the reviewers.
    \item We changed the model architecture slightly, allowing attention between the interventional and the observational data instead of only allowing information flow in one direction only. We further added a query token to ensure the mask is properly encoded and the model is positionally invariant. Instead of using a mean pooling and modelling an explicit difference between the distributions, we use an attention pooling. These changes led to performance improvements in our experiments. Additionally, these changes move the architecture further away from that of \cite{Dhir2507}, addressing a comment of the reviewers. 
We added the simple baseline of \cite{fang2026rethinking} as a comparison algorithm, as suggested by reviewer \textit{zCDN}.
    \item We removed the finetuning experiments in the main experimental section, since this was a major point of criticism by all reviewers and caused some confusion. By adapting the prior and the architecture, our results match the ones of our initially finetuned models now.
    \item We moved the multi-root cause identification experiment to the appendix. 
    \item We present the result tables for PetShop and causRCA more compactly and moved the full results to the appendix. We have run all but the CausTR+PC algorithms (specialised for CausRCA) on both data sets now and report the results. We only report the performance of graph-free algorithms in the main body now, as the joint reporting of graph-free and graph-aware algorithms caused confusion among the reviewers.
\end{enumerate}
The initial submission of this work which contains the results of the old model architecture and prior, is available on~\href{https://arxiv.org/abs/2605.08786v3}{arXiv}.

\section{Reproducibility}

We publish the code with the Paper. For all the experiments and evaluation we used a fixed random seed to ensure reproducibility.
The synthetic prior used to train PRIM, including graph sampling, structural equation families, exogenous noise, availability mode, and intervention types and calibration, is described in Section~\ref{sec:setup} and specified in full detail in Appendix~\ref{app:data-gen} and Appendix~\ref{app:prior}, with calibrated intervention-strength ranges given in Table~\ref{tab:intervention-mmd}. The padding and masking strategy used to handle variable graph sizes is described in Appendix~\ref{app:padding}. Full architectural details of PRIM, including the data encoding, the alternating-attention transformer blocks, the query-pooling decoder, and the training objective, are given in Appendix~\ref{app:architecture}. All hyperparameters for training and the experiments, the model used per experiment, and compute requirements are listed in Appendix~\ref{app:hyperparams} (Tables~\ref{tab:hyperparams}--\ref{tab:compute}). Details of the PetShop and CausRCA benchmarks, including data splits and preprocessing, are given in Section~\ref{sec:real} of the main text, with additional results in Appendices~\ref{app:pet} and~\ref{app:causrca}. For more explanations on these data sets we refer to \cite{pmlr-v236-hardt24a} for the PetShop and \cite{Mehling2026} for the CausRCA data. Baselines marked with an asterisk in our tables are numbers reported directly by the original papers; all other baseline numbers are computed by our own code, calling the original authors' implementations rather than reimplementing them where available: Traversal, CIRCA, $\varepsilon$-diagnosis, RCD, random-walk, and counterfactual attribution use the implementations bundled with the PetShop benchmark~\cite{pmlr-v236-hardt24a}; Baro uses RCAEval's~\cite{pham2025rcaeval} own implementation; TimeRank (for PetShop, where it is not reported by the original authors) uses causRCA's~\cite{Mehling2026} own time-recency baseline, adapted to PetShop's continuous telemetry via a threshold-based change-detection step we describe in code; the PC-discovered graph variants use \texttt{causal-learn}'s \citep{zheng2024causal} PC implementation; ranked-correlation and the percentile baseline are our own, graph-free implementations. Both the PetShop and CausRCA evaluations, including all baselines, can be run end-to-end on a laptop CPU with no GPU required (verified in under two minutes each), letting readers verify our reported numbers without access to a GPU cluster. We provide the code for that in the supplemental.

\section{Derivations and Identifiability of Root Causes}

\subsection{Derivation of the Bayesian Posterior}\label{app:post}

Since the components of $\mathbf{f} \equiv (\mathbf{f}^{\text{obs}}, \mathbf{f}^{\text{int}})$ are not independent, we for convenience do not model both functions separately in the posterior.
The full distribution we want to estimate is:
\begin{equation}
    p(T, \mathbf{f}, \mathcal{G}| \mathcal{D}^{\text{obs}}, \mathcal{D}^{\text{int}}, m)
\end{equation}
by marginalising out $\mathcal{G}$ we get
\begin{equation}
      p(T,\mathbf{f}| \mathcal{D}^{\text{obs}}, \mathcal{D}^{\text{int}}, m) =  \sum_\mathcal{G}  p(T, \mathbf{f}, \mathcal{G} | \mathcal{D}^{\text{obs}}, \mathcal{D}^{\text{int}}, m)
\end{equation}
by additionally marginalising out $\mathbf{f}$ we get
\begin{equation}
    p(T |  \mathcal{D}^{\text{obs}}, \mathcal{D}^{\text{int}}, m) = \int \left ( \sum_\mathcal{G}  p(T, f, \mathcal{G} | \mathcal{D}^{\text{obs}}, \mathcal{D}^{\text{int}}, m) \right) d \mathbf{f}
\end{equation}
which we can rewrite using the chain rule to
\begin{equation}
       \sum_\mathcal{G} \int p(T \mid \mathbf{f}, \mathcal{G}, \mathcal{D}^{\text{obs}}, \mathcal{D}^{\text{int}}, m)\, p(\mathbf{f} \mid \mathcal{G}, \mathcal{D}^{\text{obs}}, \mathcal{D}^{\text{int}}, m)\, p(\mathcal{G} \mid \mathcal{D}^{\text{obs}}, \mathcal{D}^{\text{int}}, m) \, d  \mathbf{ f}
\end{equation}

The factorization of the distribution of the Bayesian Model gives the the following conditional independence statements: \begin{enumerate}
    \item \textbf{Sufficiency of the mechanism, location, and graph for $T$}
    \begin{equation}
    p\big(T \mid     X^{\text{obs}},     X^{\text{int}}, \mathbf{f}, m, \mathcal{G}\big) 
    = p\big(T \mid \mathbf{f}, m, \mathcal{G}\big)
    \end{equation}
    If we have the graph $\mathcal{G}$, the mechanism change $\mathbf{f}$ and the location of the observed anomaly the data does not carry any more information about the root cause $T$.

    \item \textbf{Irrelevance of the anomaly indicator to the graph}
    \begin{equation}
    p\big(\mathcal{G} \mid     X^{\text{obs}},     X^{\text{int}}, m\big) 
    = p\big(\mathcal{G} \mid     X^{\text{obs}},     X^{\text{int}} \big)
    \end{equation}
    By definition of our Bayesian model the anomaly indicator $m$ does not carry any information about the graph; this means the graph is determined by the data alone the observed anomaly does not tell anything about $\mathcal{G}$ since it only occurs after the graph is already known/sampled.

    \item \textbf{Irrelevance of the anomaly indicator to the mechanism}
    \begin{equation}
    p\big(\mathbf{f} \mid \mathcal{G},     X^{\text{obs}},     X^{\text{int}}, m\big) 
    = p\big(\mathbf{f} \mid \mathcal{G},     X^{\text{obs}},     X^{\text{obs}}\big)
    \end{equation}
    In a similar way the functional difference $\mathbf{f}$ does not depend on where the anomaly was observed. This can intuitively be explained as the observed anomaly only happens after the functional shift/ root cause of the anomlay has occurred. 
\end{enumerate}

We can therefore rewrite this as
\begin{equation}
    p(T |  \mathcal{D}^{\text{obs}}, \mathcal{D}^{\text{int}}, m) =  \sum_{\mathcal{G}} \int p(T \mid \mathbf{f}, \mathcal{G}, m)\, p(\mathbf{f} \mid \mathcal{G}, \mathcal{D}^{\text{obs}}, \mathcal{D}^{\text{int}})\, p(\mathcal{G} \mid \mathcal{D}^{\text{obs}}, \mathcal{D}^{\text{int}}) \, d \mathbf{f}.
\end{equation}

\clearpage

\subsection{Identifiability of the Root Cause(s)}\label{app:id_proof}

Here we provide our reasoning for why the posterior $p(T \mid \mathcal{D}^{\text{obs}}, \mathcal{D}^{\text{int}}, m)$ concentrates on the correct $T$ in the infinite-data limit. The argument is fully nonparametric and covers both soft and hard interventions. We focus on the multi-sample interventional setting, in which each interventional regime provides enough samples to identify its conditional distributions; this covers all our experiments except the single-anomaly CausRCA case. The argument relies on the identifiability result of \citet{squires2020permutation}.

\paragraph{Setting} We consider $K \geq 1$ interventional regimes. For each regime $k$, let $T^k_{\text{full}}$ denote the set of nodes whose mechanism is altered in regime $k$ relative to the observational regime, and write $T_{\text{full}} = \bigcup_{k=1}^K T^k_{\text{full}}$ for the union of all mechanism-changed nodes across regimes. The root cause set $T \subseteq T_{\text{full}}$ is the subset of mechanism-changed nodes that are causally responsible for the observed anomaly $m$, as defined in the main text.

\begin{definition}[$\mathcal{I}$-Markov equivalence \citep{yang2018characterizing,squires2020permutation}]
\label{def:i-mec}
Two pairs of DAG and intervention target list, $(\mathcal{G}_1, T_{\text{full},1})$ and $(\mathcal{G}_2, T_{\text{full},2})$, are \emph{$\mathcal{I}$-Markov equivalent} if they induce the same set of observational and interventional distributions over $X$, where each interventional regime preserves the conditional $p(X_i \mid \mathrm{pa}_{\mathcal{G}}(i))$ for non-intervened nodes. The equivalence class of $(\mathcal{G}, T_{\text{full}})$ under this relation is the \emph{$\mathcal{I}$-Markov equivalence class} ($\mathcal{I}$-MEC).
\end{definition}

\begin{assumption}[Acyclic SCM]
\label{ass:dag}
The data $\mathcal{D}^{\text{obs}}$ and $\mathcal{D}^{\text{int}} = \{\mathcal{D}^{\text{int},k}\}_{k=1}^K$ are generated by a structural causal model whose causal graph $\mathcal{G}$ is a directed acyclic graph over the observed variables $X = (X_1, \dots, X_n)$.
\end{assumption}

\begin{assumption}[Causal sufficiency]
\label{ass:sufficiency}
There are no latent confounders.
\end{assumption}

\begin{assumption}[Mechanism invariance for non-targets]
\label{ass:invariance}
For each interventional regime $k$ and each $X_i \notin T^k_{\text{full}}$, the conditional $p(X_i \mid \mathrm{pa}_{\mathcal{G}}(i))$ is invariant between the observational regime and regime $k$. Both soft and hard interventions on the targets in $T^k_{\text{full}}$ are permitted.
\end{assumption}

\begin{assumption}[$\mathcal{I}$-faithfulness; \citet{squires2020permutation}, Assumption 2]
\label{ass:i-faithfulness}
The observational distribution is faithful to $\mathcal{G}$ in the usual sense, and conditional invariances across regimes correspond exactly to $d$-separations between intervention nodes and observed variables in the \emph{augmented graph} $\mathcal{G}^{\mathcal{I}}$ (the graph obtained from $\mathcal{G}$ by adding, for each regime $k$, an intervention node $\zeta_k$ with edges $\zeta_k \to X_i$ for every $X_i \in T^k_{\text{full}}$). Intuitively, this rules out interventions that leave no statistical trace: every genuine mechanism change must produce a detectable difference in some conditional distribution, and effects from distinct interventions or distinct paths must not cancel exactly.
\end{assumption}

\begin{assumption}[Anomaly variable]
\label{ass:anomaly}
The observed anomaly $m$ corresponds to a variable in $X$ whose anomalous behaviour is induced (directly or indirectly) by mechanism changes at a subset $T \subseteq T_{\text{full}}$ of intervened nodes. We refer to $T$ as the \emph{root cause set}.
\end{assumption}

\begin{proofthm}[Proof of \textbf{\autoref{thm:identifiability}}]
We first invoke the identifiability result of \citet{squires2020permutation}. Their Assumption~2 (our Assumption~\ref{ass:i-faithfulness}) states that for any regime $k$ and any disjoint $A, C$,
\[
  (A \perp\!\!\!\perp \zeta_k \mid C \cup \zeta_{[K]\setminus\{k\}})_{\mathcal{G}^{\mathcal{I}}} \quad\Longleftrightarrow\quad f^k(x_A \mid x_C) = f^{\text{obs}}(x_A \mid x_C).
\]
The forward direction is the $\mathcal{I}$-Markov property (Assumptions~\ref{ass:dag},~\ref{ass:sufficiency}); the reverse is faithfulness (Assumption~\ref{ass:i-faithfulness}), which rules out coincidental cancellations that would mask a genuine intervention.

Applying this with $A = \{i\}$ identifies the intervention targets, as in \citet[\S 3.2]{squires2020permutation}. A node $i$ lies in $T^k_{\text{full}}$ iff $\zeta_k$ and $i$ are adjacent in $\mathcal{G}^{\mathcal{I}}$, and adjacent nodes cannot be d-separated by any conditioning set. The biconditional above (Assumption~\ref{ass:i-faithfulness}) translates this into the data-determined statement that no $C$ makes $f^k(x_i \mid x_C)$ and $f^{\text{obs}}(x_i \mid x_C)$ agree; for non-targets, conditioning on $\mathrm{pa}_{\mathcal{G}}(i)$ does, by mechanism invariance (Assumption~\ref{ass:invariance}). Hence $T^k_{\text{full}}$, and so $T_{\text{full}} := \bigcup_k T^k_{\text{full}}$, is identifiable, for both soft and hard interventions (Assumption~\ref{ass:invariance}).

The set $T_{\text{full}}$ contains every node whose mechanism has changed in any regime, regardless of whether that change is causally relevant to the observed anomaly. Providing $m$ gives the following: a node $i \in T_{\text{full}}$ can only have contributed to the anomaly at $m$ if it lies upstream of $m$ in the causal graph (Assumption~\ref{ass:anomaly}). Mechanism changes at nodes that are not ancestors of $m$ propagate nothing to $m$ and are therefore irrelevant to the root cause analysis, even though they are genuine interventions. The root cause set is thus
\[
  T = \{ i \in T_{\text{full}} : i \in \mathrm{An}_{\mathcal{G}}(m) \}.
\]
The same biconditional (Assumption~\ref{ass:i-faithfulness}), applied with $A = \{m\}$, identifies $T$. Under causal sufficiency (Assumption~\ref{ass:sufficiency}), $\zeta_k$ is d-connected to $m$ exactly when some target in $T^k_{\text{full}}$ is an ancestor of $m$ in $\mathcal{G}$ (Assumption~\ref{ass:dag}), so a node in $T_{\text{full}}$ is an ancestor of $m$ if no conditioning set makes the interventional and observational distributions of $m$ agree, a pattern determined by the data alone. Every DAG in the $\mathcal{I}$-MEC must therefore agree on which nodes in $T_{\text{full}}$ are ancestors of $m$, even if it disagrees about the rest of the ancestor relation. Marginalising over the $\mathcal{I}$-MEC yields the correct $T$ in the infinite-data limit, and residual ambiguities among nodes outside $T_{\text{full}}$ do not affect $T$.
\end{proofthm}

\paragraph{Compatibility of the prior with the assumptions}
Our prior generation fulfils these assumptions by design: we sample only DAGs (no cycles) with no latent confounders, satisfying Assumptions~\ref{ass:dag} and~\ref{ass:sufficiency}. Soft interventions are modelled so as not to alter the graph structure, and hard interventions only drop the parents of the target node (more details in Appendix~\ref{app:data-gen}); in both cases the conditionals of non-target nodes are unchanged, satisfying Assumption~\ref{ass:invariance}. Because mechanism parameters are drawn from continuous distributions, the probability that an intervention is exactly cancelled by another intervention or by the functional form of a downstream mechanism is zero (a standard generic-position argument), so Assumption~\ref{ass:i-faithfulness} holds almost surely. Access to the anomaly variable is standard in RCA evaluations and is provided by our prior generation, satisfying Assumption~\ref{ass:anomaly}. Thus both the prior generation and the synthetic evaluation setups satisfy all the assumptions used above.

\clearpage

\section{Data and Prior Design}
\subsection{Data Generation}\label{app:data-gen}

\begin{figure}[h!]
    \centering
    \includegraphics[width=1\linewidth]{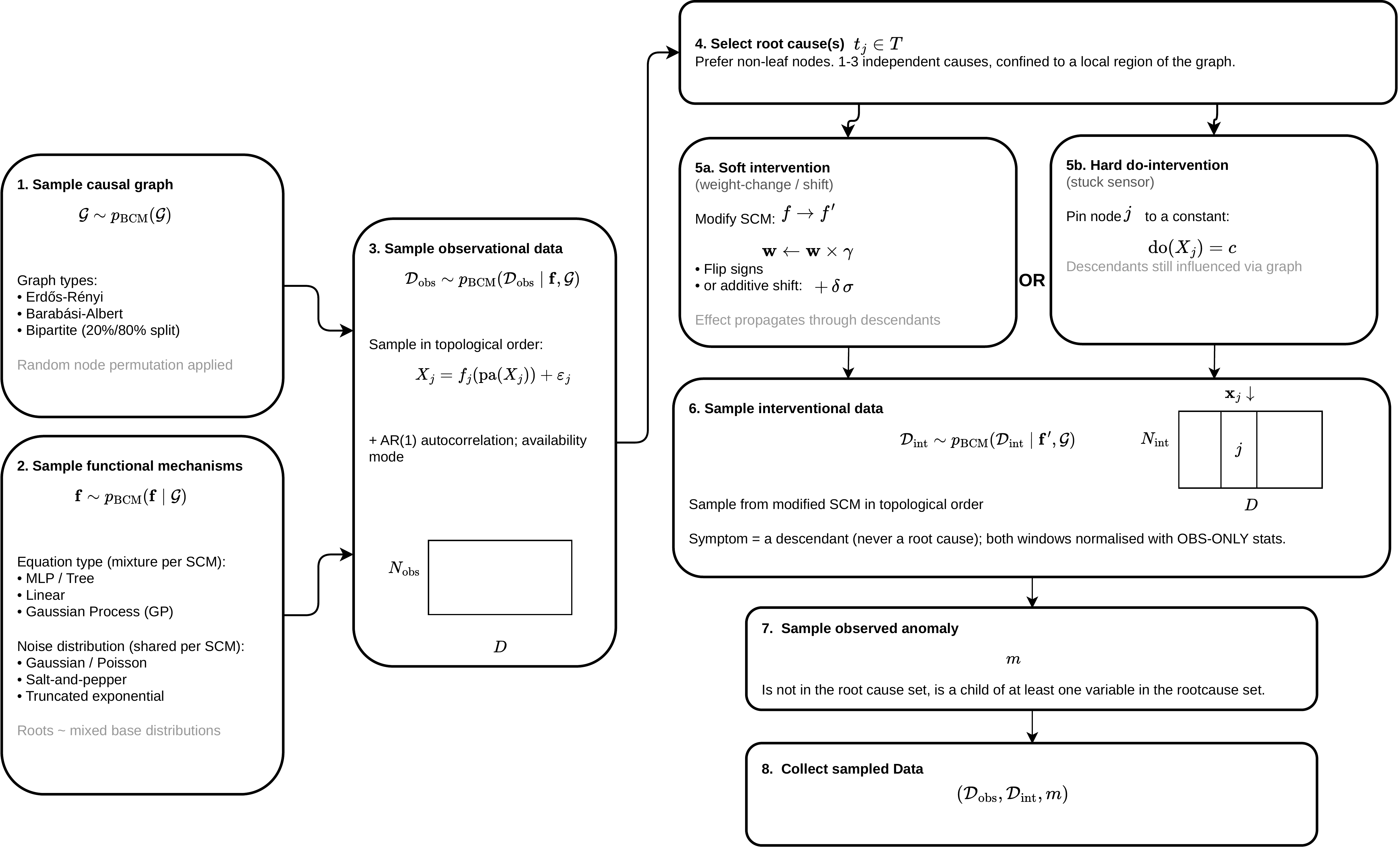}
    \caption{Overview of the data generation process. A causal graph $\mathcal{G}$
    and functional mechanisms $\mathbf{f}$ are sampled to define an SCM.
    Observational data $\mathcal{D}_{\text{obs}}$ is drawn by ancestral sampling,
    after which a target node $t_j \in T$ is selected and the SCM is perturbed
    via a soft intervention ($\mathbf{f}^{\text{obs}} \!\to \mathbf{f}^{\text{int}}$)
    or a hard do-intervention pinning $t_j$ to a constant.
    Interventional data $\mathcal{D}_{\text{int}}$ is then sampled from the
    modified SCM.}
    \label{fig:datagen}
\end{figure}

The more detailed data generation process is illustrated in Figure \ref{fig:datagen}. For each
training instance, a DAG $\mathcal{G}$ and a set of structural equations
$\mathbf{f}$ are sampled to form an SCM.

Observational data $\mathcal{D}_{\text{obs}} \in \mathbb{R}^{N_{\text{obs}} \times D}$
is collected by forward-sampling in topological order. Root-cause nodes $T$ are then selected and a soft or hard intervention is applied,
yielding interventional data
$\mathcal{D}_{\text{int}} \in \mathbb{R}^{N_{\text{int}} \times D}$.
The triple $(\mathcal{D}_{\text{obs}},\, \mathcal{D}_{\text{int}},\, T)$
forms one training example. Note for all experiments except the finetuning for multiRCA and CausRCA $|T| = 1$ meaning we only inject one root cause.

\subsection{Prior over Causal Models}\label{app:prior}
The prior design follows the design of \citet{qu2026tabiclv} with some specific changes to make it more suitable for the root cause analysis tasks such as different types of interventions, constant baselines and a rich range of noise distributions as well as only sampling sparse graphs.

\paragraph{Graph structure}
As the usual set-up for root cause analysis rarely includes dense causal graph we assume a sparse structure of the causal DAG.
Each episode samples a DAG $\mathcal{G}$ over
$K\!\sim\!\mathrm{Unif}(K_{\min},K_{\max})$ nodes.
The graph family is chosen uniformly from Erd\H{o}s--R\'enyi (ER),
Barab\'asi--Albert (BA), and bipartite.
Edge density is controlled via expected in-degree
$\bar{d}\!\sim\!\mathrm{Unif}(1.8,2.5)$, giving
$p_{\text{edge}}=\bar{d}/(K{-}1)$.
A uniform node-index permutation removes topological-ordering bias.

\paragraph{Structural equations}
An episode draws one mechanism family per SCM from \{MLP, tree, linear, GP\}, with mixture weights $(0.50, 0.13, 0.22, 0.15)$. The MLP and tree mechanisms, the root-variable sampler, and the discretisation are taken from the TabICL prior \citep{qu2026tabiclv}. 
Follwing them we initialise small, random MLPs with block-sparse weights and injected per-layer noise, and tree ensembles (decision tree / random forest / extra-trees / gradient boosting) fit to random synthetic targets, yielding piecewise-constant, threshold-style mechanisms; see \citet{qu2026tabiclv} for details. The linear mechanism is a simplified $X_k = \mathbf{w}_k^\top \tilde{X}_{\mathrm{pa}(k)} + \varepsilon_k$, and the Gaussian-process mechanism is unchanged (RBF/Mat\'ern kernel, log-uniform lengthscale, Gamma-distributed observation noise), with an auxiliary $\mathcal{N}(0,1)$ input, independently resampled per node, appended to the kernel inputs so the mechanism's output is not purely additive in the observation noise.

Root nodes are sampled independently of the mechanism family, from a per-node distribution fixed once at initialisation (Normal / Uniform / Laplace / Student-$t$ / multimodal / mixed), following TabICL's \texttt{XSampler} \citep{qu2026tabiclv}.

\paragraph{Availability mode}
With probability $0.35$, an SCM instead mimics a near-constant failure metric: every node operates near a high baseline with tiny noise.

\paragraph{Exogenous noise}
Additionally we let each SCM draws its noise distribution uniformly from \{Gaussian, Poisson, salt-and-pepper, truncated-exponential\} with a low, log-uniform scale, and with probability $0.6$ applies AR(1) autocorrelation to mimic the slowly-varying signals typical of monitoring data. This enables our prior to work with time series data.

\paragraph{Interventions/ Sampling of Root Causes}
This is to mimic the failure of a part of the system enabling the pior to be used for root cause analysis.
Each node in $T$ is intervened on independently with one of weight-change ($62\%$), additive shift ($25\%$), hard do-intervention ($8\%$), or noise inflation ($5\%$). 
Interventions act on the mechanism's \emph{output}, so they apply identically across mechanism families and to root and non-root nodes ($\mu_k,\sigma_k$: observational mean/std of $X_k$; $s\!\sim\!\mathrm{Unif}\{-1,+1\}$):
\begin{itemize}
    \item \textbf{Weight change:} $X_k' = \mu_k + s\gamma(X_k-\mu_k)$, $\gamma\!\sim\!\mathrm{Unif}(5,9)$, sign flipped w.p.\ $0.8$.
    \item \textbf{Additive shift:} $X_k' = X_k + s\delta\sigma_k$, $\delta\!\sim\!\mathrm{Unif}(1,4)$.
    \item \textbf{Hard do-intervention:} $X_k' = \mu_k + sm\sigma_k$, $m\!\sim\!\mathrm{Unif}(1,4)$, collapsing $\mathrm{Var}(X_k)$ to zero (``stuck sensor'').
    \item \textbf{Noise inflation:} the exogenous noise scale is multiplied by $c\!\sim\!\mathrm{Unif}(3,6)$.
\end{itemize}
In availability mode, shift and do-intervention instead drop the metric by a fraction of the baseline.
We then sample the observed anomli(es) $m$, which are children of the sampled root cause.

\subsection{Padding and Variable Graph Size}
\label{app:padding}

PRIM operates on graphs of varying size $k \leq K_{\max}$, where $K_{\max}$ is the maximum number of nodes the model was trained on. To handle variable-size inputs with a fixed-size transformer, we follow the padding strategy of \citet{hollmanntabpfn, Hollmann2025}.

\paragraph{Data padding} For a graph with $k$ real nodes, both $\mathcal{D}^{\text{obs}}$ and $\mathcal{D}^{\text{int}}$ are zero-padded along the node dimension to size $K_{\max}$. Real node values are z-score normalised and then scaled by a factor of $K_{\max}/k$ before padding, so that the magnitude of activations remains consistent regardless of graph size. Padded positions are filled with zero. The normalisation is based on the observational data only.

\paragraph{Attention masking} A binary padding mask is constructed such that positions $j \geq k$ are marked as padding. This mask is passed to all transformer layers, where it is used as a key padding mask in the node-level attention step, preventing padded positions from contributing to attention scores. After every LayerNorm operation, padded positions are explicitly re-zeroed by elementwise multiplication with the mask, preventing LayerNorm from reintroducing non-zero values at padding locations.

\paragraph{Output masking} At inference time, logits at padding positions $j \geq k$ are set to $-\infty$ before the softmax, ensuring the model can only predict valid nodes as root causes. During training, the same masking is applied before the cross-entropy loss.

\subsection{Parameters of the intervention Strengths}

 For each intervention type we ran an RBF-kernel MMD two-sample permutation test \citep{JMLR:v13:gretton12a} ($\alpha=0.05$, $n{=}110$ episodes per magnitude) between the observational and interventional windows, and set the lower bound of each range to the smallest tested magnitude at which the test's rejection rate is $\gtrsim 0.9$. This means the weakest sampled fault is still reliably detectable. We show the values in \autoref{tab:intervention-mmd}.

\begin{table}[h!]
\centering
\caption{Calibrated intervention-strength ranges.}
\label{tab:intervention-mmd}
\begin{tabular}{lcccl}
\toprule
Intervention & Mixture weight $\pi_k$ & Range & Reject rate @ floor & Notes \\
\midrule
Weight-change         & $0.62$ & $\gamma\sim\mathrm{Unif}(5,9)$ & $0.93$ & flip w.p.\ $0.8$ \\
Additive shift         & $0.25$ & $\delta\sim\mathrm{Unif}(1,4)$ & $0.98$ & \\
Hard do-intervention   & $0.08$ & $m\sim\mathrm{Unif}(1,4)$      & $0.96$ & \\
Noise inflation        & $0.05$ & $c\sim\mathrm{Unif}(3,6)$      & --     & not calibrated \\
\bottomrule
\end{tabular}
\end{table}

\clearpage
\section{PRIM Architecture Details}
\label{app:architecture}

PRIM (\textbf{P}rior-fitted \textbf{R}oot cause \textbf{I}dentification with \textbf{M}eta-learning), is built around an alternating node-/sample-level attention encoder together with an attention\citep{Vaswani2017} based, Neural-Process-style readout.

Our architecture is oriented on the causal-structure-induction architecture of \citet{ke2023learning}. 

However, \cite{ke2023learning} infer a full graph structure from a set of observational and interventional samples. In contrast  our problem setting is different: the model  receives both observational (normal) and interventional (anomalous) data and predicts which node was intervened on: the root cause. 
An overview of the architecture is shown in Figure \ref{fig:arch}.
Variable graph sizes are handled by zero-padding and attention masking following \citet{hollmanntabpfn, Hollmann2025}. For this we provide implementation details in Appendix~\ref{app:padding}.

\paragraph{Data Encoding}
Observational and interventional datasets $\mathcal{D}^{\text{obs}} \!\in\! \mathbb{R}^{n_{\text{obs}}\times K}$, $\mathcal{D}^{\text{int}} \!\in\! \mathbb{R}^{n_{\text{int}} \times K}$ are encoded into tensors $\mathbf{H}^{\text{obs}}, \mathbf{H}^{\text{int}} \in \mathbb{R}^{n_{(\cdot)}\times K \times d}$ whose feature dimension $d = d_v + d_n + d_t + d_a$ is partitioned into four \emph{concatenated} segments. Each scalar $x_{s,k}$ is projected by a data-type-specific linear layer and concatenated with a node-identity, a sample-type, and an anomaly segment:
\begin{equation}
    \mathbf{H}^{\text{obs}}_{s,k} = \big[\, W^{\text{obs}} x_{s,k} \;\Vert\; \mathbf{e}^{\text{node}}_k \;\Vert\; \mathbf{e}^{\text{type}}_{0} \;\Vert\; \mathbf{e}^{\text{anom}}_{a_k} \,\big], \qquad
    \mathbf{H}^{\text{int}}_{s,k} = \big[\, W^{\text{int}} x_{s,k} \;\Vert\; \mathbf{e}^{\text{node}}_k \;\Vert\; \mathbf{e}^{\text{type}}_{1} \;\Vert\; \mathbf{e}^{\text{anom}}_{a_k} \,\big],
\end{equation}
where $W^{\text{obs}}, W^{\text{int}} \in \mathbb{R}^{d_v \times 1}$ are separate learned value embeddings.
Since observational and interventional data are different distributions, each gets its own weights rather than a shared projection. 
$\mathbf{e}^{\text{node}}_k \in \mathbb{R}^{d_n}$ is a \emph{fixed}, non-learned sinusoidal positional embedding over node index \citep{Vaswani2017}. We do since node order is randomly permuted and carries no intrinsic structure. 
 $\mathbf{e}^{\text{type}}_c \in \mathbb{R}^{d_t}$ $c\in\{0,1,2\}$ are the observational and interventional/queries respectively  and $\mathbf{e}^{\text{anom}}_{a} \in \mathbb{R}^{d_a}$ ($a\in\{0,1,2\}$ is  the non-target/target/mask-not-provided $a_k = m_k$ when the binary symptom mask $m\!\in\!\{0,1\}^K$ is available, else $a_k=2$.

A learned per-node query token $\mathbf{z}^{(0)}_k = \big[\, \mathbf{q}_k \;\Vert\; \mathbf{e}^{\text{type}}_{2} \;\Vert\; \mathbf{e}^{\text{anom}}_{a_k} \,\big]$, with free parameter $\mathbf{q}_k \in \mathbb{R}^{d_v+d_n}$, is concatenated onto the sample axis as one additional value giving
$\mathbf{Z}^{(0)} = [\mathbf{H}^{\text{obs}}; \mathbf{H}^{\text{int}}; \mathbf{z}^{(0)}] \in \mathbb{R}^{N\times K \times d}$ with $N = n_{\text{obs}}+n_{\text{int}}+1$.

\paragraph{Alternating-Attention Transformer Blocks}
$L$ pre-norm transformer blocks refine $\mathbf{Z}^{(0)}$ as a single sequence: observational and interventional samples attend to each other at every layer.
(1)~node-level attention (even $\ell$): for each of the $N$ samples, attention is computed across the $K$ node positions;
(2)~sample-level attention (odd $\ell$): for each node, attention is computed jointly across all $N$ samples -- observational, interventional, and the query token.
Each attention and each MLP sublayer sits in a pre-norm residual block:
\begin{equation}
    \hat{\mathbf{Z}}^{(\ell)} = \mathbf{Z}^{(\ell-1)} + \mathrm{Attn}^{(\ell)}\!\big(\mathrm{LN}(\mathbf{Z}^{(\ell-1)})\big), \qquad
    \mathbf{Z}^{(\ell)}_{\text{pre}} = \hat{\mathbf{Z}}^{(\ell)} + \mathrm{MLP}\!\big(\mathrm{LN}(\hat{\mathbf{Z}}^{(\ell)})\big),
\end{equation}
where "+" denotes the residual (skip) connection. After each block, the type and anomaly segments are \emph{re-written} . This means they are  overwritten, not added to  with $\mathbf{e}^{\text{type}}, \mathbf{e}^{\text{anom}}$, so every token's identity stays valid.  However deep the stack is; the value/node segments are left to evolve through the residual stream:
\begin{equation}
    \mathbf{Z}^{(\ell)} = \big[\, \mathbf{Z}^{(\ell)}_{\text{pre},\,1:d_v+d_n} \;\Vert\; \mathbf{e}^{\text{type}} \;\Vert\; \mathbf{e}^{\text{anom}} \,\big].
\end{equation}

\paragraph{Decoding}
After $L$ layers a final layer norm is applied, $\bar{\mathbf{Z}} = \mathrm{LN}_f(\mathbf{Z}^{(L)})$, and the sequence is split into the embedded samples $\bar{\mathbf{H}} = \bar{\mathbf{Z}}_{1:n_{\text{obs}}+n_{\text{int}}}$ and the embedded query token $\bar{\mathbf{z}} = \bar{\mathbf{Z}}_{N}$.  Another attention readout (QueryPooling) maps the dataset directly to a  summary/embedding for each node
\begin{equation}
    \mathbf{u} = \mathrm{LN}\Big(\bar{\mathbf{z}} + \mathrm{CrossAttn}\big(\bar{\mathbf{z}}, \bar{\mathbf{H}}, \bar{\mathbf{H}}\big)\Big) \in \mathbb{R}^{K \times d}.
\end{equation}
As no absolute positional encoding is used along the sample axis, the alternating self-attention layers and the final cross-attention layer are both invariant to the ordering of the observational and interventional samples. This means that the difference  between them is  instead learned from the type segments rather than imposed as an explicit difference. A two-layer MLP decoder then maps each node's summary to a logit: $\hat{T} = c_\psi(\mathbf{u}) \in \mathbb{R}^K$.

\clearpage

\section{Hyperparameters and Compute}
\label{app:hyperparams}

We train a total of two models as some of the smaller experiments can be done with a simpler/smaller model trained with a smaller maximum number of variables size $K_{max}$.
All models use embedding dimension $d=160$, $\mathcal{l}=6$ attention layers, 8 attention heads, MLP hidden size 512, and dropout 0.1. The small model is trained on $80{,}000$ unique samples from $20{,}000$ distinct SCMs; the large model on $7{,}680{,}000$ samples from $1{,}920{,}000$ distinct SCMs. The small model is trained with a single root cause per episode ($|T|=1$); the large model is trained on the prior's default multi-root-cause mix, where the number of root causes is uniformly sampled between 1 and 3(Appendix~\ref{app:prior}), so that it also covers the multiple simultaneous root causes present in the CausRCA benchmark.
Hyperparameters for all training runs and experimenst are listed in Table~\ref{tab:hyperparams}; compute requirements are summarised in Table~\ref{tab:compute}; the model used per experiment is given in Table~\ref{tab:model-mapping}.

\begin{table}[h]
\centering
\caption{Hyperparameters for training and experiments}
\label{tab:hyperparams}
\begin{tabular}{llr}
\toprule
Component & Hyperparameter & Value \\
\midrule
\multirow{9}{*}{\shortstack[l]{Training \\ (all models)}}
  & Optimizer                   & AdamW \\
  & Learning rate               & $5\times10^{-4}$ \\
  & Weight decay                & $0.01$ \\
  & Episodes / epoch            & $1{,}000$ \\
  & Queries / episode ($n_q$)   & $4$ \\
  & $n_{\text{obs}}$            & $\mathrm{Unif}(5,\,500)$ \\
  & $n_{\text{int}}$            & $\mathrm{Unif}(1,\,200)$ \\
  & Dropout                     & $0.1$ \\
\midrule
\multirow{2}{*}{\shortstack[l]{Training \\ (per model)}}
  & Small ($K_{\min}=2,\;K_{\max}=5$),\;40 epochs   & $4\times$A100-80GB \\
  & Large ($K_{\min}=17,\;K_{\max}=100$),\;160 epochs & $8\times$A100-80GB \\
\midrule
\multirow{5}{*}{Multi-RC FT}
  & Learning rate               & $3\times10^{-5}$ \\
  & Epochs                      & $5$ \\
  & $n_{\text{obs}}$            & $\mathrm{Unif}(40,\,150)$ \\
  & $n_{\text{int}}$            & $\mathrm{Unif}(5,\,20)$ \\
\midrule
\multirow{3}{*}{Ablation FT}
  & Learning rate               & $5\times10^{-5}$ \\
  & Epochs                      & $10$ \\
  & Episodes / epoch            & $500$ \\
\bottomrule
\end{tabular}
\end{table}

\begin{table}[h]
\centering
\caption{Model used per experiment.}
\label{tab:model-mapping}
\begin{tabular}{llr}
\toprule
Experiment & Model & $K_{\max}$ \\
\midrule
Three-node confounder / mediator          & PRIM-S  & 10   \\
Finetuning ablation (sinusoidal, far OOD) & PRIM-S & 10   \\
Finetuning ablation (NN, near OOD)        & PRIM-S & 10  \\
Multi-root-cause identification           & PRIM-S & 10  \\
Scalability (20--100 nodes)               & PRIM-L  & 100 \\
PetShop                                   & PRIM-L  & 100 \\
CausRCA                                   & PRIM-L  & 100 \\
Inference time benchmark                  & PRIM-L  & 100 \\
\bottomrule
\end{tabular}
\end{table}

\makeatletter
\setlength{\@fptop}{0pt}
\setlength{\@fpsep}{8pt plus 1fil}
\setlength{\@fpbot}{0pt plus 1fil}
\makeatother

\begin{table}[t!]
\centering
\caption{Compute requirements. Epochs serve solely as a logging mechanism.}
\label{tab:compute}
\begin{tabular}{lrrrr}
\toprule
Component & Parameters & Hardware & Wall time & GPU-hrs \\
\midrule
Small model  & $1{,}928{,}409$ & $4\times$A100-80GB & 41m      & $\approx 5.5$   \\
Large model  & $1{,}939{,}209$ & $8\times$A100-80GB & 76h\,37m & $\approx 612.9$ \\
\midrule
Evaluation   & —               & $1\times$A100      & 2h       & $\approx 2.0$   \\
\midrule
\textbf{Total (reported)}      & & & & $\approx 620$  \\
\textbf{Total (incl.\ trials)} & & & & $\approx 3102$ \\
\bottomrule
\end{tabular}
\end{table}

\clearpage
\section{Additional Experiments}

\subsection{Multi-Root Cause Identification}\label{app:multirca}

\begin{figure}[h!]
    \centering
    \begin{subfigure}[c]{0.18\textwidth}
        \centering
        \begin{tikzpicture}[
            node distance=0.8cm,
            main/.style = {draw, shape=circle, minimum size=0.55cm, inner sep=0pt, font=\small},
            root/.style = {draw, shape=circle, minimum size=0.55cm, inner sep=0pt, font=\small, fill=red!70},
            desc/.style = {draw, shape=circle, minimum size=0.55cm, inner sep=0pt, font=\small, fill={rgb,255:red,255; green,179; blue,102}}
        ]
            \node[root] (X1)                              {$X_1$};
            \node[root] (X2) [below=0.7cm of X1]         {$X_2$};
            \node[desc] (X3) [right=0.9cm of X1]         {$X_3$};
            \node[desc] (X4) [right=0.9cm of X2]         {$X_4$};
            \node[main] (X5) [above=0.6cm of X3]         {$X_5$};
            \node[desc] (X6) [right=0.9cm of X3, yshift=-0.35cm] {$X_6$};
            \draw[->] (X1) -- (X3);
            \draw[->] (X2) -- (X3);
            \draw[->] (X2) -- (X4);
            \draw[->] (X3) -- (X6);
            \draw[->] (X4) -- (X6);
            \draw[->] (X5) -- (X6);
        \end{tikzpicture}
    \end{subfigure}%
    \hspace{0.03\textwidth}
    \begin{subfigure}[c]{0.5\textwidth}
        \centering
        \includegraphics[width=0.8\textwidth]{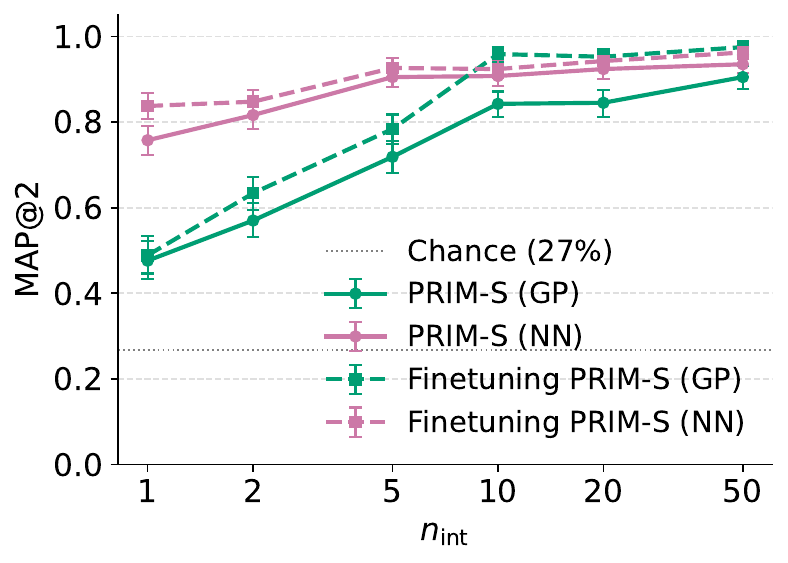}
        \caption{Varying $n_{\mathrm{int}}$, keeping $n_{\text{obs}} = 100$}
    \end{subfigure}
    \caption{Multi-root-cause evaluation on a 6-node DAG (left).
    $X_1, X_2$ ({\color{red!70}$\blacksquare$} red) are root causes;
    $X_3, X_4, X_6$ ({\color{rgb,255:red,255;green,179;blue,102}$\blacksquare$} orange)
    are descendants; $X_5$ (white) is unaffected.
    MAP@2 measures whether both appear in the top-2 predictions.
    Solid: zero-shot; dashed: finetuned on multi-fault DAGs.}
    \label{fig:multi_rca}
\end{figure}

Here PRIM's ability to identify multiple simultaneous root causes on the fixed DAG shown in Figure~\ref{fig:multi_rca} is tested, where two independent nodes $X_1$ and $X_2$ are intervened upon simultaneously. Performance is measured by MAP@2, requiring both root causes to appear in the top-2 predictions, testing zero-shot transfer from single to multi-fault identification. The fine-tuned variant is trained on 1,600 synthetic multi-fault scenarios with mixed NN/GP mechanisms and full-model fine-tuning; details are given in Appendix~\ref{app:hyperparams}.

As shown in Figure~\ref{fig:multi_rca} for a fixed $n_{\text{obs}} = 100$, the fine-tuned variant substantially outperforms the base model, with gains becoming more pronounced as $n_{\mathrm{int}}$ increases and GP mechanisms are handled more reliably. Notably, even the base model exceeds the random MAP@2 baseline of ${\approx}0.27$ with sufficient interventional samples, suggesting that single-fault training transfers partially to the multi-fault setting.

\clearpage
\subsection{Finetuning and out of Distribution Generalisation}\label{app:finetune}
\label{app:ablation_ft}

We evaluate the benefit of finetuning and the choice of finetuning variants across two out-of-distribution settings. This also showcases how PRIM adapts to out of distribution data.

In both cases we use random DAGs with 3–10 nodes and sweep over the number of interventional samples $n_{\mathrm{int}}$, holding $n_{\mathrm{obs}}=100$ fixed.
In both settings finetuning uses AdamW with learning rate $5\times10^{-5}$, weight decay 0.01, 10 epochs of 500 episodes each, and $n_{\text{obs}}=100$ fixed. They both use $k_{max} = 10$

\emph{Sinusoidal} (far OOD): structural equations of the form $f(x)=a\sin(bx+c)$ with randomly sampled parameters, entirely absent from the meta-training prior, and an intervention strength of $2\times$ (vs.\ $3\times$ during training).
\emph{NN equations} (near OOD): the same NN equation family as training but with a reduced intervention strength of $1.5\times$ (vs.\ $3\times$ during training), weakening the $\boldsymbol{\Delta}$ signal while keeping the functional form in-distribution.
Three conditions are compared: zero-shot (no finetuning), decoder-only finetuning (backbone frozen), and full-model finetuning (all parameters updated). Finetuning in both cases uses synthetic data drawn from the respective target distribution.

\begin{figure}[h!]
\centering
\begin{subfigure}{0.45\textwidth}
\centering
\includegraphics[width=\linewidth]{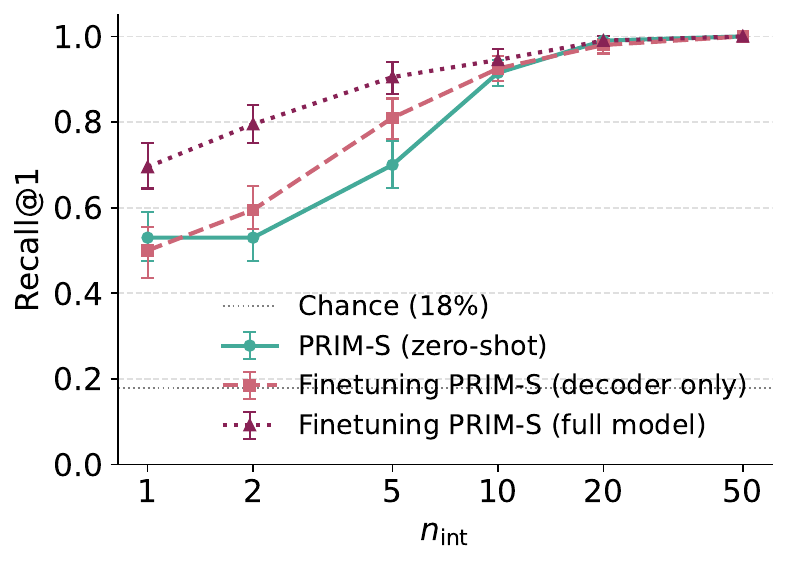}
\caption{Sinusoidal equations (far OOD)}
\label{fig:ablation_sin}
\end{subfigure}%
\hspace{0.05\textwidth}
\begin{subfigure}{0.45\textwidth}
\centering
\includegraphics[width=\linewidth]{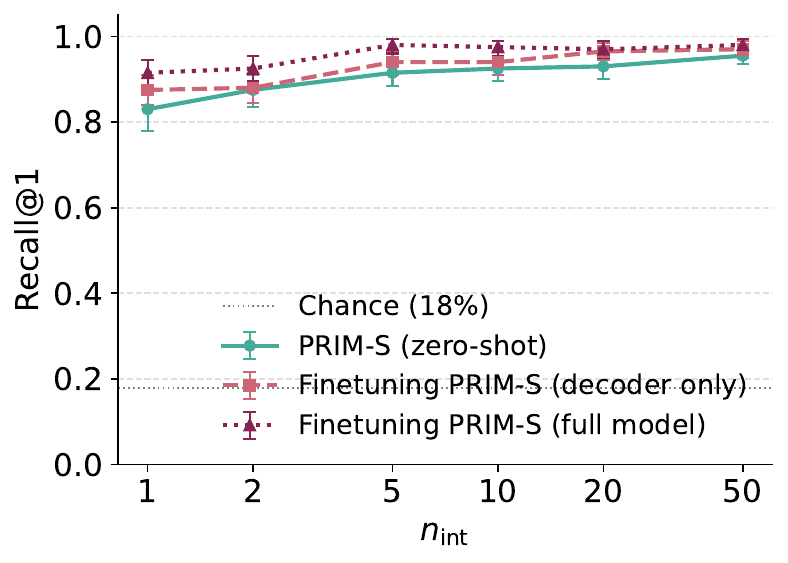}
\caption{Weak-signal NN equations (near OOD)}
\label{fig:ablation_weak}
\end{subfigure}
\vspace{-0.5em}
\caption{Recall@1 vs.\ $n_{\mathrm{int}}$ at fixed $n_{\mathrm{obs}}=100$ for three finetuning conditions. Finetuning uses synthetic data from the target distribution. Error Bars show 90\% bootstrap confidence intervals.}
\label{fig:ablation_ft}
\end{figure}

As shown in Figure~\ref{fig:ablation_ft}, finetuning consistently reduces the number of interventional samples needed to achieve high recall, with the largest gains at small $n_{\mathrm{int}}$.
For the far-OOD sinusoidal setting (Figure~\ref{fig:ablation_sin}), all three conditions are clearly separated: full-model finetuning achieves ${\approx}0.70$ recall at $n_{\mathrm{int}}=1$ versus ${\approx}0.45$ for zero-shot, with decoder-only falling between the two. This indicates that the backbone itself needs to adapt to the novel functional form, and updating all parameters is beneficial.
For the near-OOD NN setting (Figure~\ref{fig:ablation_weak}), zero-shot performance is already high (${\approx}0.83$ at $n_{\mathrm{int}}=1$) despite the reduced intervention strength, and decoder-only finetuning nearly matches full-model finetuning across the entire sweep. This is consistent with the backbone already producing well-calibrated embeddings for the in-distribution functional form.
However, with sufficient interventional data PRIM also converges to the true root causes even when the posterior is out of its prior distribution.

\clearpage
\clearpage
\subsection{Additional Results Different Node Sizes}\label{app:diff_nodes}
\begin{figure}[h!]
    \centering
    \includegraphics[width=\textwidth]{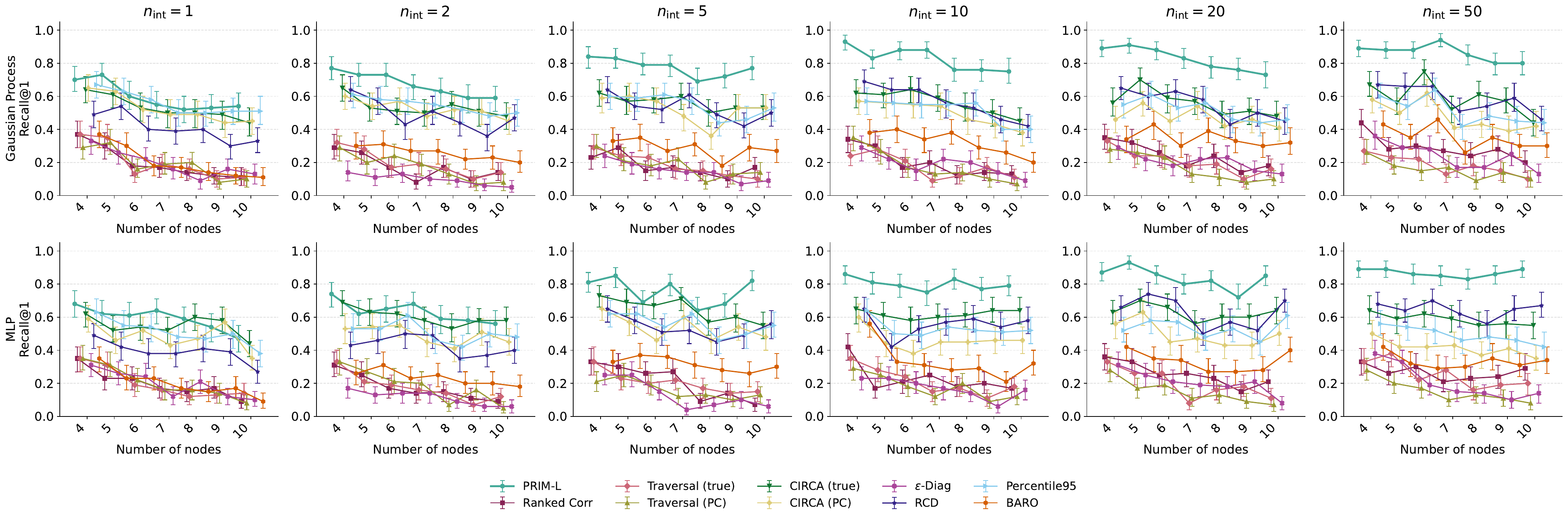}
    \caption{Recall@1 vs.\ number of nodes across all observation sizes
    ($n_{\mathrm{obs}} \in \{50, 100, 200, 500\}$, $n_{\mathrm{int}}=10$).
    Rows correspond to data generating process (Gaussian Process and Neural Network);
    columns correspond to number of observations. For smaller Model (3 to 10  Nodes)}
    \label{fig:errorbar_all_10}
\end{figure}

\begin{figure}[h!]
    \centering
    \includegraphics[width=\textwidth]{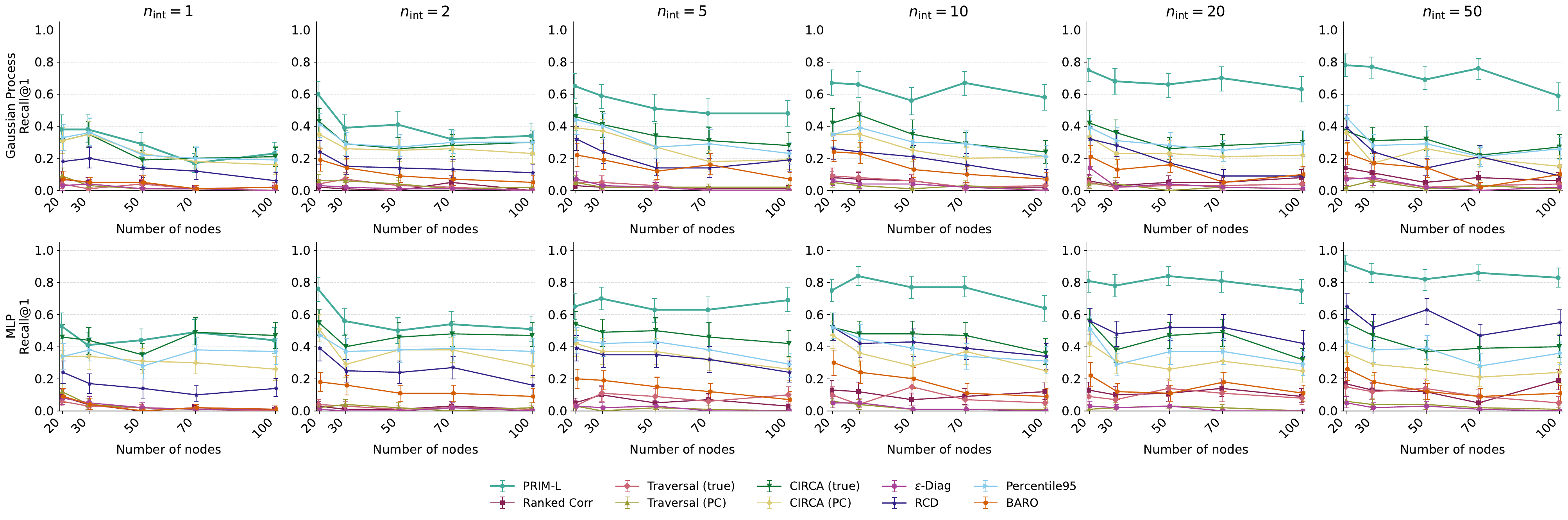}
    \caption{Recall@1 vs.\ number of nodes across all observation sizes
    ($n_{\mathrm{obs}} \in \{50, 100, 200, 500\}$, $n_{\mathrm{int}}=10$).
    Rows correspond to data generating process (Gaussian Process and Neural Network);
    columns correspond to number of observations. For large Model trained on (17 - 100 Nodes)}
    \label{fig:errorbar_all}
\end{figure}

\clearpage
\subsection{Inference Time}\label{app:time}
\begin{figure}[h!]
    \centering
    \includegraphics[width=0.5\linewidth]{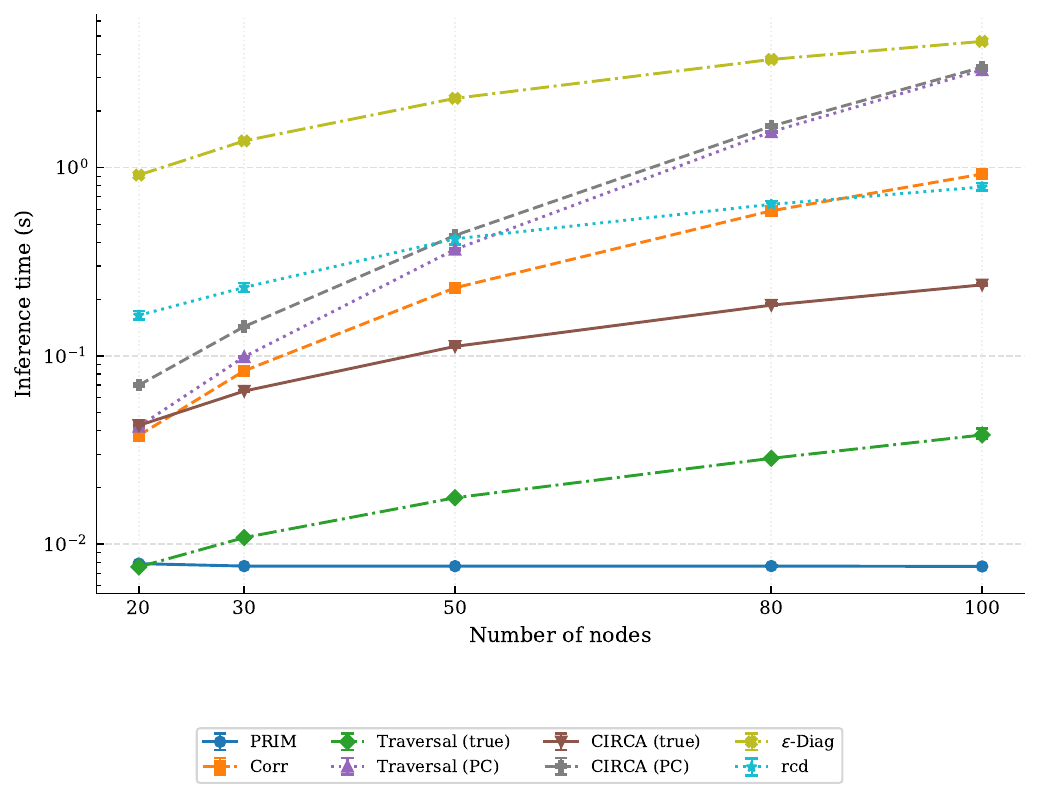}
    \caption{Inference time for different node sizes for different methods $n_{\text{obs}}=100$, $n_{\text{int}}=20$}
    \label{fig:placeholder}
\end{figure}

\begin{table}[h]
\centering
\caption{PRIM inference time on a MacBook. Mean and 90\% bootstrap CI over 100 episodes per configuration ($n_{\text{obs}}=100$, $n_{\text{int}}=20$).}
\label{tab:inference_laptop}
\begin{tabular}{lrrrr}
\toprule
Device & $K$ & Mean (ms) & CI$_{\text{lo}}$ (ms) & CI$_{\text{hi}}$ (ms) \\
\midrule
\multirow{5}{*}{CPU} 
  & 20  & 115.8 & 115.2 & 116.5 \\
  & 30  & 117.2 & 116.8 & 117.6 \\
  & 50  & 119.7 & 119.3 & 120.2 \\
  & 80  & 122.6 & 122.2 & 123.0 \\
  & 100 & 124.5 & 124.1 & 125.0 \\
\midrule
\multirow{5}{*}{MPS (Apple GPU)}
  & 20  & 17.4 & 16.7 & 18.5 \\
  & 30  & 16.8 & 16.8 & 16.9 \\
  & 50  & 16.6 & 16.6 & 16.7 \\
  & 80  & 16.5 & 16.5 & 16.6 \\
  & 100 & 16.5 & 16.4 & 16.5 \\
\bottomrule
\end{tabular}
\end{table}

\clearpage

\subsection{Additional Results For Petshop} \label{app:pet}
\begin{table*}[h!]
\caption{Top-1 recall of the RCA methods measuring the accuracy of identifying the correct root-cause node. Results marked with * are taken from \citet{pmlr-v236-hardt24a}; $\dagger$ denotes our method. \textbf{Bold} indicates the best result per row overall; \textcolor{blue}{blue} indicates the best result per row among graph-not-given methods.}
\label{tab:top1recall}
\centering
\resizebox{\textwidth}{!}{%
\begin{tabular}{ll|ccc|cccccc}
\toprule
& & \multicolumn{3}{c|}{graph given} & \multicolumn{6}{c}{graph not given} \\
\cmidrule(lr){3-5} \cmidrule(lr){6-11}
traffic & metric & traversal\textsuperscript{*} & circa\textsuperscript{*} & counter- & $\varepsilon$-diag.\textsuperscript{*} & rcd\textsuperscript{*} & corr\textsuperscript{*} & Baro & TimeRank & PRIM$^\dagger$ \\
scenario & & & & factual\textsuperscript{*} & & & & & & \\
\midrule
low      & latency      & \textbf{0.57} & 0.36 & 0.36 & 0.00 & 0.07 & \textcolor{blue}{0.43} & 0.00 & 0.00 & 0.30 \\
low      & availability & 0.50 & 0.42 & 0.00 & 0.00 & 0.58 & \textbf{\textcolor{blue}{0.75}} & 0.00 & 0.00 & 0.63 \\
high     & latency      & 0.57 & 0.50 & 0.57 & 0.00 & 0.00 & \textbf{\textcolor{blue}{0.64}} & 0.00 & 0.00 & 0.50 \\
high     & availability & 0.33 & 0.00 & 0.00 & 0.00 & 0.00 & \textbf{\textcolor{blue}{0.83}} & 0.00 & 0.00 & 0.38 \\
temporal & latency      & \textbf{1.00} & 0.75 & 0.38 & 0.12 & 0.25 & \textcolor{blue}{0.62} & 0.00 & 0.00 & 0.33 \\
temporal & availability & 0.38 & 0.38 & 0.00 & 0.00 & 0.50 & 0.62 & 0.00 & 0.00 & \textbf{\textcolor{blue}{0.67}} \\
\midrule
\multicolumn{2}{l|}{average} & 0.56 & 0.40 & 0.22 & 0.02 & 0.23 & \textbf{\textcolor{blue}{0.65}} & 0.00 & 0.00 & 0.46 \\
\bottomrule
\end{tabular}%
}
\end{table*}

\begin{table*}[h!]
\caption{Top-3 recall of the RCA methods for the PetShop data measuring the accuracy of including the correct root-cause node in the top-3 results. Results marked with * are taken from \citet{pmlr-v236-hardt24a}; $\dagger$ denotes our method. \textbf{Bold} indicates the best result per row overall; \textcolor{blue}{blue} indicates the best result per row among graph-not-given methods.}
\label{tab:top3recall}
\centering
\resizebox{\textwidth}{!}{%
\begin{tabular}{ll|ccc|cccccc}
\toprule
& & \multicolumn{3}{c|}{graph given} & \multicolumn{6}{c}{graph not given} \\
\cmidrule(lr){3-5} \cmidrule(lr){6-11}
traffic & metric & traversal\textsuperscript{*} & circa\textsuperscript{*} & counter- & $\varepsilon$-diag.\textsuperscript{*} & rcd\textsuperscript{*} & corr\textsuperscript{*} & Baro & TimeRank & PRIM$^\dagger$ \\
scenario & & & & factual\textsuperscript{*} & & & & & & \\
\midrule
low      & latency      & 0.57 & \textbf{0.86} & 0.71 & 0.00 & 0.21 & 0.57 & 0.30 & 0.10 & \textcolor{blue}{0.80} \\
low      & availability & \textbf{1.00} & \textbf{1.00} & 0.42 & 0.00 & 0.75 & \textcolor{blue}{0.92} & 0.00 & 0.00 & 0.88 \\
high     & latency      & 0.79 & \textbf{1.00} & 0.86 & 0.00 & 0.07 & \textcolor{blue}{0.79} & 0.00 & 0.30 & 0.60 \\
high     & availability & \textbf{1.00} & 0.00 & 0.00 & 0.33 & 0.00 & \textcolor{blue}{0.92} & 0.00 & 0.00 & 0.88 \\
temporal & latency      & \textbf{1.00} & \textbf{1.00} & 0.50 & 0.12 & 0.75 & \textcolor{blue}{0.75} & 0.17 & 0.17 & 0.67 \\
temporal & availability & \textbf{1.00} & \textbf{1.00} & 0.25 & 0.12 & 0.75 & 0.75 & 0.00 & 0.33 & \textbf{\textcolor{blue}{1.00}} \\
\midrule
\multicolumn{2}{l|}{average} & \textbf{0.86} & 0.81 & 0.46 & 0.10 & 0.39 & 0.78 & 0.08 & 0.15 & \textcolor{blue}{0.79} \\
\bottomrule
\end{tabular}%
}
\end{table*}

\subsection{Additional Results for CausRCA}\label{app:causrca}

\begin{table}[h!]
\centering
\caption{MAP@3 for RCA algorithms on the CausRCA benchmark. Results marked with * are taken from \citet{Mehling2026}; $^\dagger$Ours. \textbf{Bold} indicates best result per row overall; \textcolor{blue}{blue} indicates best result per row among graph-not-given methods.}\label{tab:causrca_res}
\resizebox{\textwidth}{!}{%
\begin{tabular}{ll|ccccc|ccccccc}
\toprule
& & \multicolumn{5}{c|}{graph given} & \multicolumn{7}{c}{graph not given} \\
\cmidrule(lr){3-7}\cmidrule(lr){8-14}
Dataset & & CausTR\textsuperscript{*} & Rand & Page & Traversal & CIRCA & Time & Baro\textsuperscript{*} & Corr & CausTR & $\varepsilon$-Diag & RCD & PRIM$^{\dagger}$ \\
        & &                          & Walk\textsuperscript{*} & Rank\textsuperscript{*} & & & Rank\textsuperscript{*} & & & +PC\textsuperscript{*} & & & \\
\midrule
Coolant    & Full & \textbf{0.99} & 0.00 & 0.00 & 0.00 & 0.20 & 0.35 & 0.00 & 0.00 & -- & 0.00 & 0.01 & \textcolor{blue}{0.457} \\
           & Sub  & \textbf{1.00} & 0.14 & \textbf{1.00} & 0.00 & 0.21 & 0.85 & \textcolor{blue}{0.88} & 0.15 & 0.87 & 0.00 & 0.41 & 0.760 \\
Hydraulics & Full & \textbf{0.74} & 0.00 & 0.11 & 0.00 & 0.17 & 0.18 & 0.26 & 0.13 & -- & 0.00 & 0.08 & \textcolor{blue}{0.313} \\
           & Sub  & 0.74 & 0.14 & \textbf{0.98} & 0.00 & 0.20 & 0.47 & \textcolor{blue}{0.75} & 0.35 & 0.48 & 0.02 & 0.21 & 0.745 \\
Probe      & Full & \textbf{0.57} & 0.00 & 0.00 & 0.00 & 0.01 & \textcolor{blue}{0.19} & 0.00 & 0.00 & -- & 0.06 & 0.01 & 0.086 \\
           & Sub  & 0.58 & 0.32 & 0.26 & 0.00 & 0.23 & 0.47 & \textcolor{blue}{\textbf{0.63}} & 0.45 & 0.37 & 0.29 & 0.14 & 0.353 \\
\midrule
average & Full & \textbf{0.77} & 0.00 & 0.04 & 0.00 & 0.12 & 0.24 & 0.09 & 0.05 & -- & 0.02 & 0.04 & \textcolor{blue}{0.272} \\
           & Sub  & \textbf{0.78} & 0.20 & 0.75 & 0.00 & 0.21 & 0.60 & \textcolor{blue}{0.75} & 0.33 & 0.57 & 0.11 & 0.24 & 0.616 \\
\bottomrule
\end{tabular}}
\end{table}

\clearpage

\end{document}